\documentclass[pdflatex,sn-mathphys-num]{sn-jnl}

\usepackage{graphicx}%
\usepackage{multirow}%
\usepackage{amsmath,amssymb,amsfonts}%
\usepackage{amsthm}%
\usepackage{mathrsfs}%
\usepackage[title]{appendix}%
\usepackage[table]{xcolor}%
\usepackage{textcomp}%
\usepackage{manyfoot}%
\usepackage{booktabs}%
\usepackage{algorithm}%
\usepackage{algorithmicx}%
\usepackage{algpseudocode}%
\usepackage{listings}%
\usepackage{hyperref}
\usepackage{array}
\usepackage{soul}
\usepackage[capitalise]{cleveref}
\usepackage{xcolor}

\usepackage{verbatim}

\everypar\expandafter{\the\everypar\looseness=-1}
\usepackage{xspace}
\newcommand{\ModelName}{\textsc{CEMRAG}\xspace}

\definecolor{incorrect}{RGB}{255,200,200}   
\definecolor{incomplete}{RGB}{255,230,180}  
\definecolor{superfluous}{RGB}{255,255,200} 
\definecolor{accurate}{RGB}{200,255,200}    

\newcommand{\hlincorrect}[1]{{\sethlcolor{incorrect}\hl{#1}}}
\newcommand{\hlincomplete}[1]{{\sethlcolor{incomplete}\hl{#1}}}
\newcommand{\hlsuperfluous}[1]{{\sethlcolor{superfluous}\hl{#1}}}
\newcommand{\hlaccurate}[1]{{\sethlcolor{accurate}\hl{#1}}}

\title[Article Title]{Concept-Enhanced Multimodal RAG: Towards Interpretable and Accurate Radiology Report Generation}

\author[1]{\fnm{Marco} \sur{Salmè}}\email{marco.salme@unicampus.it}

\author[2]{\fnm{Federico} \sur{Siciliano}}\email{siciliano@diag.uniroma1.it}

\author[2]{\fnm{Fabrizio} \sur{Silvestri}}\email{fsilvestri@diag.uniroma1.it}

\author*[1,3]{\fnm{Paolo} \sur{Soda}}\email{paolo.soda@umu.se}

\author[4]{\fnm{Rosa} \sur{Sicilia}}\email{rosa.sicilia@unicamillus.org}
\equalcont{These authors contributed equally to this work.}

\author[1]{\fnm{Valerio} \sur{Guarrasi}}\email{valerio.guarrasi@unicampus.it}
\equalcont{These authors contributed equally to this work.}

\affil[1]{\orgdiv{Department of Engineering}, \orgname{Research Unit of Artificial Intelligence and Computer Systems, Università Campus Bio-Medico of Roma}, \city{Rome}, \country{Italy}}

\affil[2]{\orgdiv{Department of Computer, Control and Management Engineering}, \orgname{Sapienza University of Rome}, \city{Rome},  \country{Italy}}

\affil[3]{\orgdiv{Department of Diagnostics and Intervention, Radiation Physics, Biomedical Engineering}, \orgname{Umeå University}, \city{Umeå},  \country{Sweden}}

\affil[4]{\orgname{UniCamillus-Saint Camillus International University of Health Sciences}, \city{Rome}, \country{Italy}}

\makeatletter
\renewcommand\paragraph{\@startsection{paragraph}{4}{\z@}%
  {1ex \@plus 1ex \@minus .2ex}%
  {-1em}%
  {\normalfont\normalsize\bfseries}}
\makeatother

\begin{document}

\abstract{
Radiology Report Generation (RRG) through Vision-Language Models (VLMs) promises to reduce documentation burden, improve reporting consistency, and accelerate clinical workflows.
However, their clinical adoption remains limited by the lack of interpretability and the tendency to hallucinate findings misaligned with imaging evidence.
Existing research typically treats interpretability and accuracy as separate objectives, with concept-based explainability techniques focusing primarily on transparency, while Retrieval-Augmented Generation (RAG) methods targeting factual grounding through external retrieval.
We present Concept-Enhanced Multimodal RAG (\ModelName), a unified framework that decomposes visual representations into interpretable clinical concepts and integrates them with multimodal RAG. 
This approach exploits enriched contextual prompts for RRG, improving both interpretability and factual accuracy.
Experiments on MIMIC-CXR and IU X-Ray across multiple VLM architectures, training regimes, and retrieval configurations demonstrate consistent improvements over both conventional RAG and concept-only baselines on clinical accuracy metrics and standard NLP measures. 
These results challenge the assumed trade-off between interpretability and performance, showing that transparent visual concepts can enhance rather than compromise diagnostic accuracy in medical VLMs.
Our modular design decomposes interpretability into visual transparency and structured language model conditioning, providing a principled pathway toward clinically trustworthy AI-assisted radiology. The project page is available at \href{https://github.com/marcosal30/cemrag-rrg}{https://github.com/marcosal30/cemrag-rrg}.
}

\keywords{Radiology Report Generation, Vision-Language Models, Medical Imaging, Interpretability, Retrieval-Augmented Generation, Multimodal AI}

\maketitle

\section{Introduction}\label{sec1}
Vision-Language Models (VLMs)~\cite{zhang2024vision} have emerged as a breakthrough technology in medical imaging.
By jointly modeling images and textual data, they have demonstrated remarkable capabilities across several clinical applications, including visual question answering, image classification, disease diagnosis, and automated report generation~\cite{van2024large,hartsock2024vision}. 
Among these applications, Radiology Report Generation (RRG) presents a particularly challenging task: generating comprehensive textual reports from medical images that accurately describe imaging findings and identify potential pathologies. 
In this domain, VLMs offer the potential to streamline radiological workflows by automating the initial drafting of reports.
However, despite these advantages, the adoption of VLMs in clinical settings is limited by two critical factors.
Firstly, VLMs lack interpretability~\cite{zhao2024explainability}, operating as black boxes that do not reveal how visual evidence observed in medical images translates into diagnostic statements within generated reports.
Without visibility into the anatomical structures or radiological patterns that support specific diagnostic statements, clinicians cannot verify the model’s reasoning, undermining both clinical trust and patient safety.
Secondly, VLMs are prone to hallucinations \cite{huang2025survey}, producing medically inaccurate statements that are misaligned with imaging findings, such as reporting non-existent pathologies, incorrect anatomical localizations, or impressions inconsistent with observed abnormalities. 
Such inaccuracies pose particular concerns in radiology, where the recognition of subtle findings and the precise alignment between visual evidence and domain-specific terminology are essential for reliable diagnostic support.

Current research efforts have predominantly tackled these issues independently.
Existing interpretability approaches~\cite{vatsa2024adventures} provide useful insights but often operate as post-hoc explanations that do not meaningfully influence the model’s predictions.
Recently, techniques such as Sparse Linear Concept Embeddings (SpLiCE)~\cite{bhalla2024interpreting} have shown that visual embeddings can be decomposed into sparse, human-interpretable concepts without requiring manual annotations, offering a scalable pathway toward transparency in VLMs.
In parallel, recent work aimed at improving factual grounding has increasingly relied on Retrieval-Augmented Generation (RAG)~\cite{gao2023retrieval}.
By retrieving similar cases and reports from a database, RAG provides external context to ground outputs in existing knowledge, reducing hallucinations and improving clinical relevance in RRG~\cite{bernardi2024report, sun2024fact}. 
Yet, RAG-based models remain limited by retrieval errors, as the context may be insufficient, noisy, redundant or irrelevant, leading to diluting the model’s focus, potential misattribution, and even factual inconsistencies~\cite{yu2024evaluation}.

These two research areas, interpretability and factual grounding, have evolved largely in isolation.
This separation is reinforced by the widespread assumption that transparency and performance trade off against each other, by the empirical tendency of deeper, more accurate models to be harder to interpret, and by the practical difficulty of building systems that are both highly accurate and transparently reasoned~\cite{ennab2024enhancing}.
We challenge this assumption by asking: \textit{Can interpretable visual concepts be integrated into retrieval-augmented report generation to jointly improve transparency and factual accuracy in medical VLMs?}

To answer this question, we present Concept-Enhanced Multimodal RAG (\ModelName), a unified framework that reconciles interpretability and factual accuracy in RRG by integrating interpretable visual concept extraction with multimodal RAG. 
The core innovation of our approach lies in transforming interpretable visual concepts from passive post-hoc explanations into active components of the generation pipeline, using them to prioritize clinically pertinent portions of retrieved content and direct the model toward information supported by visual evidence.
Our experimental framework spans two established radiology benchmarks, MIMIC-CXR and IU-Xray, which differ substantially in scale and report characteristics. 
We examine both in-domain retrieval scenarios, where similar cases are retrieved from the same dataset, and cross-domain retrieval settings, where reports are retrieved from an external database, reflecting realistic clinical deployments where knowledge bases may originate from different institutional sources. For each retrieval configuration, we evaluate two distinct VLM architectures under two training paradigms: a Zero-Shot prompting setting that assesses the immediate effectiveness of \ModelName without model adaptation, and a Supervised Fine-Tuning (SFT) regime that examines how interpretable concepts interact with task-specific optimization. 

Our main contributions are summarized as follows:
\begin{itemize}
    \item We propose \ModelName, a framework that integrates interpretable visual decomposition with retrieval-based grounding to enhance transparency and factual accuracy in RRG.
    \item We provide the first systematic comparison of RAG and SFT paradigms in RRG, establishing a comprehensive benchmark that evaluates their individual and combined effectiveness across two VLM architectures, two retrieval configurations, and two radiology datasets.
    \item We demonstrate that \ModelName consistently outperforms traditional RAG and concept-based approaches on both Natural Language Processing (NLP) and clinical accuracy metrics across diverse experimental conditions.
    \item We provide empirical evidence that interpretable visual concepts can enhance rather than compromise factual accuracy, challenging the assumption of a trade-off between interpretability and performance in medical AI systems.
\end{itemize}

The remainder of this paper is organized as follows. We begin by reviewing related work on interpretability for VLMs and multimodal RAG in \cref{sec2}, then present our proposed framework in \cref{sec:method}. \cref{sec4} describes the experimental setup, including datasets, baselines, and evaluation metrics. We present both quantitative and qualitative results in \cref{sec:results}, and conclude with discussion and future research directions in \cref{sec6}.

\section{Related Work}\label{sec2}
Our work builds upon two complementary research areas that we review in turn. 
First, we discuss interpretability methods for VLMs, then we examine the application of multimodal RAG to medical domains.

\subsection{Interpretability for Vision-Language Models}

The interpretability of VLMs has emerged as a critical requirement for clinical deployment, particularly in medical imaging where diagnostic decisions directly impact patient outcomes. 
However, most interpretability research focuses on classification tasks, leaving the integration of transparency mechanisms into generative frameworks like RRG largely unexplored.

Current interpretability methods can be distinguished by whether they rely on implicit explanation mechanisms or explicit concept representations.
In the first category, post-hoc textual explanation techniques, exemplified by rationale generation~\cite{park2018multimodal,nguyen2024langxai} and Chain-of-Thought reasoning~\cite{wei2022chain,zheng2023ddcot}, train models to articulate their reasoning processes after making predictions through natural language. 
While these methods have demonstrated improvements in perceived transparency and, in some cases, task performance, they often act as plausible rationalizations rather than faithful reflections of the underlying computational mechanisms~\cite{chen2024measuring}. 
This limitation is particularly concerning in medical settings, where clinicians require insight into the actual diagnostic cues driving the model’s predictions.

Mechanistic Interpretability~\cite{bereskamechanistic,sharkey2025open,jiang2025devils} represents a more ambitious approach, seeking to reverse-engineer neural network computations through detailed analysis of attention patterns, information flow, and component functionality~\cite{huo2024mmneuron,conmy2023towards}. 
Despite its theoretical appeal in pursuing causal rather than correlational understanding, the architectural complexity of modern large-scale VLMs renders this approach extremely challenging to implement systematically, especially in real-world medical AI systems.

An alternative paradigm achieves interpretability through explicit concept representations.
Concept Bottleneck Models~\cite{koh2020concept,rao2024discover} exemplify this approach by forcing all information to flow through an intermediate layer of predefined human-interpretable concepts, ensuring that predictions depend exclusively on explicit semantic attributes. 
This architectural constraint offers genuine transparency but comes at a substantial cost: it demands extensive manual concept annotation and restricts model expressiveness, limiting the ability to capture visual patterns that fall outside the predefined space.

Recent work on representation decomposition offers a more flexible concept-based approach by decomposing dense embeddings into interpretable components without requiring manual annotations or architectural constraints.  
Methods such as those proposed by~\cite{gandelsmaninterpreting,balasubramanian2024decomposing} leverage internal model components such as attention mechanisms to translate high-dimensional representations into natural language descriptions, revealing how models encode visual features and their relationships to semantic concepts.
However, methods relying on attention-based decomposition remain dependent on the learned representations of these internal mechanisms, which may not align with domain-specific interpretability requirements in specialized contexts where transparency demands explicit grounding in established vocabulary.
Recently, an emerging line of work has pursued interpretability by explicitly leveraging predefined, domain-specific vocabularies of clinically meaningful concepts, enabling explanations to be expressed directly in human-readable terms rather than inferred post hoc from latent attention patterns~\cite{parekh2024concept, bhalla2024interpreting}.
Among these, SpLiCE~\cite{bhalla2024interpreting} operationalizes this paradigm by factorizing visual representations into sparse linear combinations of interpretable concepts drawn from a domain-specific vocabulary, achieving scalable and transparent explanations without sacrificing representational flexibility.

\subsection{Multimodal Retrieval-Augmented Generation in Medicine}

Multimodal RAG has emerged as a promising approach to mitigating factual hallucinations in medical VLMs by grounding generation in existing clinical knowledge. These systems extend traditional text-based retrieval by employing specialized encoders to extract features from multiple modalities, retrieving relevant cases from curated medical databases, and conditioning generation on both the input data and the retrieved context~\cite{he2025retrieval,abootorabi2025ask}. This shift from purely parametric knowledge to retrieval-augmented pipelines has yielded substantial gains across applications such as orthopedic diagnosis~\cite{jin2024orthodoc}, lung cancer staging~\cite{tozuka2025application}, and prescription interpretation for medication management~\cite{thetbanthad2025application}.

In the specific domain of RRG, multimodal RAG approaches have shown particularly promising results. 
The MMed-RAG system reports substantial improvements in factual accuracy for both medical VQA and report generation across radiology, pathology, and ophthalmology~\cite{xiammed}. Similarly, the RULE framework achieves notable gains by combining calibrated retrieval strategies with preference-based fine-tuning to balance reliance on parametric knowledge versus retrieved context~\cite{xia2024rule}. These results suggest that providing models with concrete clinical examples can substantially reduce hallucinations and improve diagnostic consistency.

Although multimodal RAG provides indirect interpretability by exposing which cases inform generation, this transparency remains fundamentally passive, revealing available information without guaranteeing how it is utilized during generation. 
Retrieval operates through global similarity matching in learned embedding spaces, capturing overall visual resemblance without explicit guidance on which anatomical structures or pathological patterns should be prioritized~\cite{wang2025mira}. 
This absence of semantic grounding creates a fundamental dilemma: insufficient retrieval fails to capture necessary clinical information, while excessive retrieval introduces irrelevant details that interfere with coherent generation~\cite{xia2024rule}. 
Consequently, models may inappropriately incorporate findings from retrieved cases that do not correspond to visual evidence in the input image, leading to factual hallucinations even in retrieval-augmented systems~\cite{chu2025reducing,sloan2024automated}.

\subsection{Limitations and Motivations}
The limitations identified across interpretability and RAG approaches reveal a fundamental gap: existing methods treat transparency and factual accuracy as separate objectives rather than mutually reinforcing components. 
Interpretability techniques provide insight into model representations without actively constraining generation toward factually grounded outputs, while RAG systems improve factual grounding without semantic control to align retrieved information with specific visual evidence. 
This separation motivates our central hypothesis that interpretable visual concepts can serve as semantic guidance mechanisms, simultaneously enhancing transparency and factual accuracy by directing retrieval and generation toward clinically relevant content present in the input image.

\section{Methodology}\label{sec:method}

To jointly improve transparency and factual accuracy in RRG, we propose \ModelName, a framework that integrates interpretable visual 
concepts with multimodal RAG through a structured prompting strategy. 
We extract clinically meaningful concepts from visual embeddings and retrieve similar cases from a database, both derived from 
the same image representation.
Rather than treating these as independent 
augmentations, we structure them hierarchically in a unified prompt that guides the language component to focus on relevant portions of retrieved context.

\begin{figure}[h]
    \centering
    \includegraphics[width=1\textwidth]{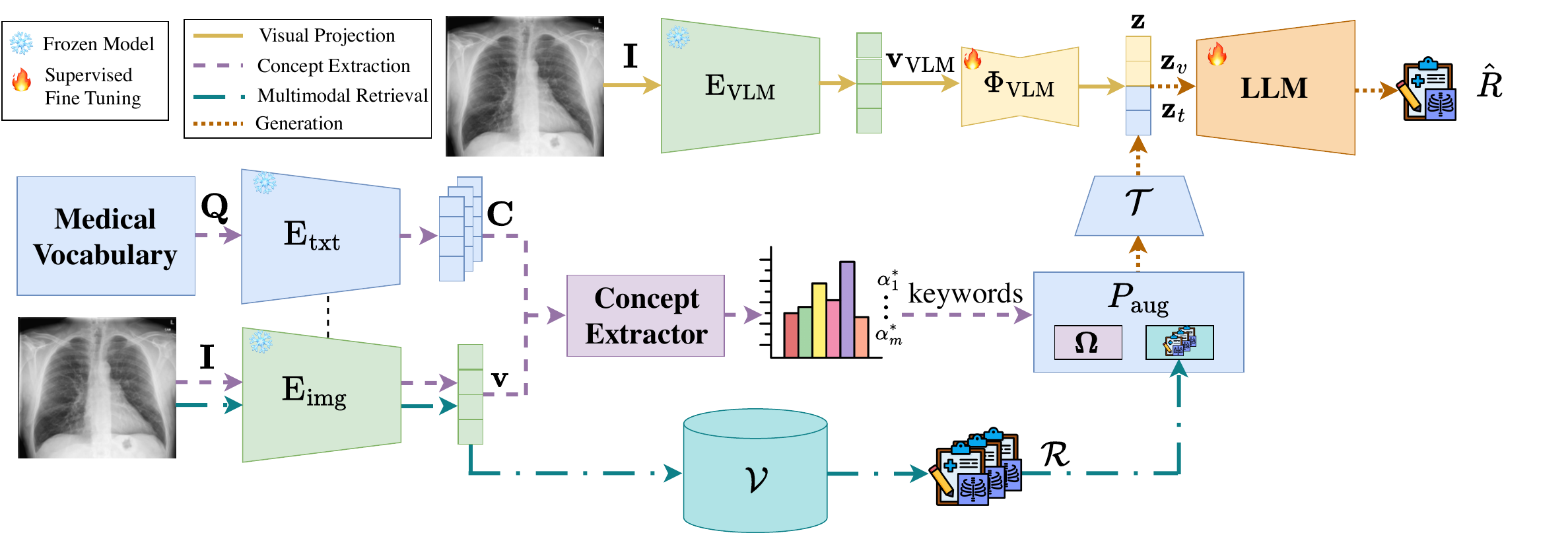}
    \caption{\ModelName overall framework, combining interpretable concept extraction with RAG for transparent and accurate radiology reporting.}
    \label{fig:framework}
\end{figure}

\cref{fig:framework} illustrates the complete architecture of our proposed framework.
Given as input a medical image $\mathbf{I} \in \mathbb{R}^{H \times W}$, where $H$ and $W$ denote the height and width in pixels, our objective is to generate a radiology report $\hat{R} = \{\hat{r}_1, \hat{r}_2, \dots, \hat{r}_{\hat{n}}\}$ that approximates the ground truth report $R = \{r_1, r_2, \dots, r_n\}$, with $r_i$ and $\hat{r}_i$ representing individual tokens, and $n$ and $\hat{n}$ denoting the number of tokens in each report. 
To achieve this, \ModelName generates reports by coordinating four key components, each represented by distinct pathways in \cref{fig:framework}:
(a) a visual encoding branch (yellow line), comprising a medical VLM encoder and a projector, extracts dense visual features from the input image;
(b) a concept-extraction module (purple dashed line) maps image embeddings into interpretable clinical keywords;
(c) a multimodal retrieval module (teal dash-dotted line) identifies similar training reports based on visual similarity in the embedding space; and
(d) an LLM (orange dotted line) conditioned on hierarchically structured prompts integrates visual tokens, concept keywords, and retrieved reports to generate the final radiology report.

\subsection{Visual Encoding and Projection}
The foundation of our framework is the extraction of dense visual features to condition the generation process.
The input image $\mathbf{I}$ is processed by a pretrained medical VLM encoder $\mathrm{E}_{\text{VLM}}$ to produce a sequence of visual features
$\mathbf{v}_{\text{VLM}} = \mathrm{E}_{\text{VLM}}(\mathbf{I}) \in \mathbb{R}^{\ell_v \times d_{\text{VLM}}}$,
where $\ell_v$ is the number of visual tokens and $d_{\text{VLM}}$ is the dimensionality of the VLM feature space.
A projection module $\Phi_{\text{VLM}}$, implemented as a token-wise multi-layer perceptron, maps the visual features into the LLM token embedding space, producing visual token embeddings
$\mathbf{z}_v = \Phi_{\text{VLM}}(\mathbf{v}_{\text{VLM}}) \in \mathbb{R}^{\ell_v \times d_{\text{LLM}}}$.
While $\mathbf{z}_v$ provides the primary visual features for generation, we employ additional aligned vision and text encoders ($\mathrm{E}_{\text{img}}$, $\mathrm{E}_{\text{txt}}$) to supply structured contextual information in a shared vision-language space, enabling both the decomposition of visual content into explicit clinical concepts and the retrieval of similar cases.

\subsection{Concept Extraction}\label{sec:concept_extraction}

The input image $\mathbf{I}$ is processed by the vision encoder $\mathrm{E}_{\text{img}}$ to produce a visual embedding $\mathbf{v} = \mathrm{E}_{\text{img}}(\mathbf{I}) \in \mathbb{R}^d$, where $d$ is the dimensionality of a shared vision-language embedding space.
To enable concept extraction, we define a medical vocabulary 
$\mathbf{Q} = \{q_1, q_2, \dots, q_m\}$
composed of $m$ domain-specific concept terms derived from the training corpus. 
Each concept $q_j \in \mathbf{Q}$ is encoded with the text encoder $\mathrm{E}_{\text{txt}}$ to obtain its embedding
$\mathbf{c}_j = \mathrm{E}_{\text{txt}}(q_j) \in \mathbb{R}^d$,
and we collect these embeddings into the concept matrix $\mathbf{C} = [\mathbf{c}_1, \mathbf{c}_2, \dots, \mathbf{c}_m] \in \mathbb{R}^{d \times m}$,
where $d$ denotes the dimensionality of the multimodal latent space.
Then, the concept extraction module decomposes the visual embedding $\mathbf{v}$ as a non-negative linear combination of concept embeddings, as detailed in Appendix~\ref{secA}.
This decomposition yields a coefficient vector $\boldsymbol{\alpha}^* \in \mathbb{R}_{\ge 0}^m$ that quantifies the 
contribution of each vocabulary concept to the image representation.
We select the top-$\tau$ concepts according to their coefficient magnitudes 
to form the interpretable keyword set $\boldsymbol{\Omega}
= [\, q_1, \dots, q_\tau \,]
\subseteq \mathbf{Q}$.

In our implementation, we instantiate $\mathrm{E}_{\text{img}}$ and $\mathrm{E}_{\text{txt}}$ using CLIP encoders~\cite{radford2021learning}, which provide aligned vision-language embeddings through contrastive pretraining. 
The same embedding $\mathbf{v}$ also enables retrieval of similar documented cases, as described next.

\subsection{Multimodal Retrieval Augmented Generation}\label{sec:MultimodalRAG}

While $\boldsymbol{\Omega}$ provides explicit clinical concepts, complete documented cases are required to ground generation in established clinical patterns and linguistic structure.
We use $\mathrm{E}_{\text{img}}$ to construct a vector database $\mathcal{V} = \{(\mathbf{v}_i^{\text{train}}, R_i^{\text{train}})\}_{i=1}^{N}$ that indexes all $N$ training images through their  visual embeddings $\mathbf{v}_i^{\text{train}}$ and associates each with its corresponding radiology report $R_i^{\text{train}}$.

Using $\mathbf{v}$ as query, the retrieval mechanism identifies the top-$k$ most similar cases by computing 
the cosine similarity $\mathcal{S}$ in the image embedding space:
\begin{equation*}
\mathcal{S}(\mathbf{I}) = \text{top-}k \left\{ 
\frac{\mathbf{v} \cdot \mathbf{v}_i^{\text{train}}}
{\| \mathbf{v} \| \| \mathbf{v}_i^{\text{train}} \|} 
\right\}_{i=1}^{N}
\end{equation*}
yielding retrieved reports $\mathcal{R} = \{ R_{i_1}^{\text{train}}, \dots, R_{i_k}^{\text{train}}\}$.

By operating in the image embedding space, this retrieval strategy prioritizes cases with similar radiological appearances, capturing visual patterns that are indicative of comparable clinical findings~\cite{you2023cxr}.

\subsection{Hierarchical Prompt Construction and Report Generation}\label{sec:prompt}
To synthesize these information sources effectively, we employ a hierarchical prompting strategy designed to mitigate the limitations of each component.
While retrieved reports $\mathcal{R}$ provide rich clinical context, they may inadvertently introduce findings absent in the query image.
Conversely, concept keywords $\boldsymbol{\Omega}$ act as precise visual anchors but lack narrative structure.
Therefore, we construct a structured prompt $P_{\text{aug}}$ where $\boldsymbol{\Omega}$ serves as a priority filter, guiding the LLM to selectively leverage the linguistic patterns in $\mathcal{R}$ that align with the observed features.

The prompt structure consists of four components: (i) a coordination instruction that establishes the task of generating a report while prioritizing concept-related content from retrieved examples, (ii) an explicit list of extracted concept keywords $\boldsymbol{\Omega}$ framed as visual findings identified in the image, and (iii) the retrieved radiology reports $\mathcal{R}$ presented as reference examples from similar cases, and (iv) a final instruction reinforcing the generation objective.
This structured presentation provides the LLM with both explicit concept annotations and implicit contextual knowledge, while maintaining a clear separation between the different information sources. 

The augmented prompt $P_{\text{aug}}$ is then processed by the LLM tokenizer $\mathcal{T}$ to produce a sequence of textual token embeddings
$\mathbf{z}_t = \mathcal{T}(P_{\text{aug}}) \in \mathbb{R}^{\ell_t \times d_{\text{LLM}}}$,
where $\ell_t$ denotes the number of text tokens and $d_{\text{LLM}}$ is the LLM embedding dimension.
The visual and textual token sequences are then concatenated along the sequence dimension to form the full multimodal input $\mathbf{z} = [\mathbf{z}_v; \mathbf{z}_t] \in \mathbb{R}^{(\ell_v + \ell_t) \times d_{\text{LLM}}}$, which is fed to the LLM.
We denote the generated report as
$\hat{R} = \mathrm{LLM}(\mathbf{z})$, and model generation in an autoregressive formulation, where the LLM predicts each token $\hat{r}_i$ conditioned on all previous tokens and on the multimodal context encoded in $\mathbf{z}$:
\[
p(\hat{R} \mid \mathbf{I}, P_{\text{aug}}) = \prod_{i=1}^{\hat{n}} p(\hat{r}_i \mid \hat{r}_{<i}, \mathbf{z}).
\]

To preserve robust medical pretrained representations, we consistently keep all the encoders frozen.
In contrast, the training strategy for the LLM and the projection module $\Phi_{\text{VLM}}$ varies according to the model configuration, as detailed in~\cref{modelConfig}.

\section{Experimental Setup}\label{sec4}
This section outlines the experimental setup, detailing the datasets, model configurations, experimental conditions, and evaluation metrics used to assess report generation quality.

\subsection{Datasets}
We conduct experiments on two well-established benchmark datasets for RRG: MIMIC-CXR~\cite{johnson2019mimic} and IU X-ray~\cite{demner2015preparing}. 
Both datasets provide chest radiographs paired with corresponding clinical reports, enabling comprehensive evaluation of our proposed approach across different data scales and clinical contexts.
The MIMIC-CXR dataset~\cite{johnson2019mimic} is a large-scale publicly available collection comprising over 370,000 chest radiographs from more than 65,000 patients.
We adopt the official training, validation, and test split, restricting our experiments to the 156,344 frontal views (posteroanterior and anteroposterior projections) due to computational limitations.
The IU X-ray dataset~\cite{demner2015preparing} consists of 7,470 chest radiographs paired with 3,955 radiological reports. This dataset serves as a complementary evaluation benchmark to assess model performance under more constrained data conditions. 
The original IU X-ray dataset includes both frontal and lateral radiological images for most reports. However, to maintain consistency with our experimental protocol on MIMIC-CXR, we only consider the 3,307 frontal projections in our experiments. 
Following established conventions in the literature~\cite{chen2021cross,liu2021exploring}, we exclude samples lacking a findings section, as this section provides the essential ground truth for supervised learning in RRG tasks. 
We employ a dataset split allocating 80\% of the data for training, 10\% for validation, and 10\% for testing, with strict enforcement of patient-level separation to prevent data leakage across splits.

\subsection{Model Configurations and Experimental Conditions}\label{modelConfig}

Our framework requires a model with aligned multimodal embeddings ($\mathrm{E}_{\text{img}},\mathrm{E}_{\text{txt}}$) capable of supporting both concept extraction and similarity-based retrieval. To this end, we employ CXR-CLIP~\cite{you2023cxr} across all experiments, which integrates a SwinTransformer~\cite{liu2021swin} as visual encoder $\mathrm{E}_{\text{img}}$ and a BioClinicalBERT~\cite{alsentzer2019publicly} as text encoder $\mathrm{E}_{\text{txt}}$, both pretrained on the MIMIC-CXR dataset.
We instantiate the concept-extraction module with Sparse Linear Concept Embeddings (SpLiCE)~\cite{bhalla2024interpreting}, which performs explicit sparse factorization of visual embeddings.
Unlike attention-based approaches, which derive concepts from learned internal representations, SpLiCE operates on a predefined medical vocabulary. This ensures that the extracted concepts are clinically relevant, and allows each term's contribution to be quantified directly through optimized sparse coefficients.
For experiments on the IU X-ray dataset, we apply Low-Rank Adaptation (LoRA)~\cite{hu2022lora} to CXR-CLIP for obtaining refined embeddings tailored to this specific data distribution (more details are given in Appendix~\ref{app:training}).

We evaluate \ModelName under two architectural configurations to assess whether the hierarchical prompting strategy remains effective across different levels of medical domain adaptation and vision-to-language alignment. 
Both configurations adopt a LLaVA-style architecture~\cite{liu2023visual} with Mistral-7B~\cite{Jiang2023Mistral7} as the LLM backbone. 
The first configuration employs LLaVA-Med~\cite{li2023llava}, where 
both the vision encoder $\mathrm{E}_{\text{VLM}}$ and the LLM incorporate medical domain-specific pretraining.
The second configuration uses CXR-CLIP as a unified encoder: $\mathrm{E}_{\text{img}}$ extracts embeddings $\mathbf{v}$ that enable concept extraction $\boldsymbol{\Omega}$, retrieval of reports $\mathcal{R}$, and generation of visual tokens $\mathbf{v}_{\text{VLM}}$.
This configuration pairs medically-pretrained CXR-CLIP with base Mistral-7B without medical language pretraining.
Unlike LLaVA-Med, which provides a pretrained projection layer, this configuration requires the introduction of a projection module $\Phi_{\text{CLIP}}$ to map CLIP visual features into the LLM embedding space.
Images are preprocessed according to each encoder's specifications: $224 \times 224$ pixels with normalization to $[-1, 1]$ for CXR-CLIP, and $336 \times 336$ center crops with encoder-specific normalization for LLaVA-Med.
Implementation details, including the initial alignment phase for $\Phi_{\text{CLIP}}$, are provided in Appendix~\ref{app:training}.

We evaluate four distinct prompting strategies that progressively incorporate retrieval and concept information. To ensure reproducibility, we provide detailed prompt templates and configuration details for each strategy in \cref{tab:experimental_conditions}.

\paragraph*{Image-Only.} This approach provides the LLM with visual features alongside a minimal prompt instructing the model to describe the radiological image. This provides a reference point for the level of performance achievable from visual information alone, without any external context. 

\paragraph*{Concepts.} This strategy augments the prompt with interpretable clinical concepts extracted from the input image using SpLiCE.
For each dataset, we construct a domain-specific vocabulary of medical concepts by selecting the 200 most frequent bigrams from the training reports.
Each bigram is encoded with the CLIP text encoder $\mathrm{E}_{\text{txt}}$ to obtain its embedding. 
SpLiCE is then applied to derive a sparse, non-negative decomposition of the image embedding over this concept vocabulary, yielding a coefficient vector that quantifies the contribution of each concept.
For each image, we rank concepts according to their coefficients, select the top five, and inject these bigrams into the prompt as explicit keywords.
Additional details regarding vocabulary construction, normalization procedures, and sensitivity analyses for different vocabulary sizes and sparsity levels are provided in the Appendix~\ref{secA1}.

\paragraph*{Multimodal RAG.} This strategy provides contextual grounding by retrieving reports from visually similar cases. 
We use FAISS~\cite{douze2024faiss} to compute cosine similarity between the query image embedding $\mathbf{v}$ derived from CXR-CLIP and embeddings in database $\mathcal{V}$, retrieving the top-$k=3$ most similar cases whose reports are then incorporated into the prompt as reference examples. 
The retrieval configuration varies by dataset.
For the MIMIC-CXR dataset, retrieval is performed within the training set of the same dataset to ensure that similar cases are drawn from the same distribution.
For IU X-ray, which has a substantially smaller training set, in-domain retrieval would provide insufficient contextual variety and limit the benefits of the RAG strategy.
We therefore implement cross-domain retrieval from the MIMIC-CXR training set, simulating a realistic scenario where a smaller institutional dataset leverages a larger external knowledge base to enhance report generation quality.

\paragraph*{\ModelName.} This method combines the two previous approaches to construct an enriched prompt that presents the top five concept bigrams as priority keywords alongside three retrieved reports. 
We design the prompt structure following Mistral's prompt engineering guidelines~\cite{mistral_prompt_guide}, which emphasize the importance of clear task definition, hierarchical organization, explicit formatting, and concrete examples for effective instruction-based generation.
Unlike previous conditions with simple generation instructions, \ModelName employs a coordination directive that establishes the task objective of generating a report from similar findings while prioritizing content related to the extracted concepts. 
Following this directive, the extracted keywords are presented in a dedicated formatted section. 
The three retrieved reports are then incorporated with separators.
The prompt concludes with a final instruction that explicitly directs the model to produce a radiology report relying on the provided findings.

This design ensures clear attribution of information sources and guides the LLM toward clinically relevant content by orienting the model toward portions of retrieved reports that align with observed visual features.

\begin{table*}[t]
\centering
\caption{Experimental configurations for radiology report generation, comparing input modalities, prompt structures, interpretability mechanisms, and factual grounding strategies. Notation: $\mathbf{v}$ denotes visual embeddings, $\boldsymbol{\Omega}$ extracted concepts and $\mathcal{R}$ retrieved reports. \texttt{Instruction}: ``Provide a description of the findings in the radiology image''; \texttt{Task}: ``Write the report of the radiology image taking information from similar FINDINGS. Consider as more relevant sentences that contain any of the KEYWORDS in the FINDINGS''; \texttt{Final Instruction}: ``Write a paragraph with only the report relying in detail on the FINDINGS''.}
\label{tab:experimental_conditions}
\resizebox{\textwidth}{!}{
\begin{tabular}{|p{2.2cm}|p{1.7cm}|p{3.8cm}|p{3.5cm}|p{3.5cm}|}
\hline
\rowcolor{gray!20}
\textbf{Strategy} & \textbf{Inputs} & \textbf{Prompt Structure} & \textbf{Interpretability} & \textbf{Factual Grounding} \\
\hline
\textbf{Image-Only} & $\mathbf{v}$ & \texttt{<Instruction>} & $\times$ black-box & $\times$ Visual encoder and LLM priors only \\
\hline
\textbf{Concepts} & $\mathbf{v}$, $\boldsymbol{\Omega}$ & \texttt{<Instruction>} + \texttt{<Keywords>}  & $\checkmark$ Concept-level via visual-semantic alignment & $\sim$ Sparse visual concept decomposition \\
\hline
\textbf{RAG} & $\mathbf{v}$, $\mathcal{R}$ & \texttt{<Instruction>} + \texttt{<Similar Findings>}& $\sim$ Indirect via retrieved context & $\checkmark$ Visual similarity-based nearest neighbours \\
\hline
\textbf{\ModelName} & $\mathbf{v}$, $\boldsymbol{\Omega}$, $\mathcal{R}$ & \texttt{<Task>} + \texttt{<Keywords>}  + \texttt{<Similar Findings>} + \texttt{<Final Instruction>}& $\checkmark\checkmark$ Dual: concept annotations and similar cases & $\checkmark\checkmark$ Concept-guided focus to salient findings\\
\hline
\end{tabular}
}
\end{table*}

Each of the four prompting strategies is evaluated under two LLM training paradigms.
In the \emph{Zero-Shot} setting, the entire model (visual encoders, projection layers, and the LLM) is kept frozen, assessing transfer capability with only prompt content varying across conditions. 
In the \emph{Supervised Fine-Tuning (SFT)} setting, we adapt the LLM using LoRA~\cite{hu2022lora} and jointly fine-tune the corresponding projection layer, while keeping all visual encoders frozen. 
This design follows the LLaVA training paradigm, isolating adaptation to the language and projection components. 
This enables a controlled investigation into how prompting strategies influence RRG performance without confounding changes in visual feature extractors.
Additional training details for both the \emph{Zero-Shot} and \emph{SFT} settings are provided in Appendix~\ref{app:training}.

For all experiments, RRG is performed with greedy decoding (temperature $T=0$) to ensure deterministic and reproducible outputs, thereby eliminating variability arising from stochastic sampling. 
All experiments are performed on four NVIDIA A100 GPUs.
On MIMIC-CXR, \emph{SFT} training time varies by strategy: Image-Only requires approximately 7 hours, \emph{Concepts} strategy 8 hours, RAG 13 hours, and \ModelName 14 hours.

\subsection{Evaluation Metrics}
Our evaluation framework incorporates two complementary categories of metrics: NLP metrics, which measure lexical similarity to reference reports, and clinical metrics, which assess the factual correctness of medical content.
To evaluate lexical similarity, we report three standard NLP metrics that are commonly used in text generation tasks. 
ROUGE-L~\cite{lin2004rouge} measures the longest common subsequence between the generated and reference texts, capturing sentence-level structural similarity. BLEU-1 and BLEU-4~\cite{papineni2002bleu} evaluate n-gram overlap at the unigram and 4-gram levels, respectively, with BLEU-4 providing assessment of longer phrasal matches that better reflect fluency and coherence. 
While these metrics effectively evaluate linguistic properties such as fluency and phrasal coherence, they capture surface-level textual similarity rather than clinical accuracy. 

To address this limitation, we emphasize clinical evaluation through two established factual correctness metrics: F1-CheXbert~\cite{zhang2020optimizing} and F1-RadGraph~\cite{delbrouck2022improving}. 
The former computes the F1 score between disease labels extracted from the generated and reference reports using the CheXbert labeler~\cite{smit2020combining}, which is a BERT-based model trained to identify the presence, absence, or uncertainty of pathological findings. Following standard practice, we report F1-CheXbert across all 14 CheXbert classes, encompassing the comprehensive range of conditions identifiable in chest radiographs.
Additionally, we report performance on the five most prevalent clinical findings in real-world chest radiograph reports: atelectasis, cardiomegaly, consolidation, edema and pleural effusion. This focused evaluation on common pathologies provides insight into the model's performance in frequently encountered diagnostic scenarios.
The second clinical metric, F1-RadGraph, quantifies factual correctness by measuring the overlap between the semantic graphs extracted from the generated and reference reports. 
This captures not only the presence of clinical findings, but also their anatomical locations and relationships between entities.
Together, these two clinical metrics provide a complementary perspective on factual correctness: F1-CheXbert focuses on diagnostic label accuracy, while F1-RadGraph assesses the representation of structured clinical content.

\section{Results}\label{sec:results}

We organize our findings into two complementary parts. 
\cref{QuantResults} presents a quantitative evaluation across datasets, model configurations, and training paradigms, using both NLP and clinical accuracy metrics. 
\cref{QualResults} provides qualitative analyses that illustrate how concept extraction and retrieval affect the generated reports and their interpretability.


\subsection{Quantitative Results}\label{QuantResults}
We structure our quantitative analysis by first examining performance on the MIMIC-CXR dataset in \cref{MimicResults}, followed by results on the IU X-ray dataset in \cref{IUxrayResults}.
For each dataset, we evaluate two model configurations: LLaVA-Med with its standard visual encoder, and LLaVA with CXR-CLIP as the visual encoder. 
Each configuration is assessed under both \emph{Zero-Shot} and \emph{Supervised Fine-Tuning (SFT)} settings to determine how domain adaptation influences the effectiveness of our approach.

\subsubsection{MIMIC-CXR}\label{MimicResults}
\cref{tab:mimic_results} presents comprehensive quantitative results on the MIMIC-CXR dataset, where retrieval is performed from the MIMIC-CXR training set, establishing an in-domain retrieval scenario.

\begin{table*}[h]
\centering
\caption{Quantitative results on the MIMIC-CXR test set for two model configurations
(LLaVA-Med and LLaVA with CXR-CLIP) and two training regimes
(\textit{Zero-Shot} and \textit{Supervised Fine-Tuning}).
We report F1-RadGraph (F1-RG), and NLP metrics (BLEU-1 as B-1, BLEU-4 as B-4, ROUGE-L as R-L). CheXbert-based label metrics are reported over 14 labels (Micro-F1$_{14}$, Macro-F1$_{14}$) and over the 5 most prevalent findings (Micro-F1$_{5}$, Macro-F1$_{5}$).
Retrieval is performed from the MIMIC-CXR training set.}
\label{tab:mimic_results}
\resizebox{\textwidth}{!}{
\begin{tabular}{llcccccccc}
\toprule
\textbf{Model} & \textbf{Method} & \textbf{Micro-F1}$_{14}$ & \textbf{Micro-F1}$_{5}$ & \textbf{Macro-F1}$_{14}$ & \textbf{Macro-F1}$_{5}$ & \textbf{F1-RG} & \textbf{B-1} & \textbf{B-4} & \textbf{R-L} \\
\midrule
\multicolumn{10}{c}{\textit{Zero-Shot}} \\
\midrule
\multirow{4}{*}{LLaVA-Med} 
& Image-Only & 0.255 & 0.237 & 0.113 & 0.134 & 0.052 & 18.81 & 0.71 & 0.131 \\
& + Concepts & 0.401 & 0.412 & 0.238 & 0.355 & 0.073 & 21.47 & 1.29 & 0.146 \\
& + RAG & 0.498 & 0.528 & 0.314 & 0.447 & 0.184 & 27.08 & 4.60 & 0.187 \\
& + \ModelName & \textbf{0.502} & \textbf{0.529} & \textbf{0.319} & \textbf{0.449} & \textbf{0.185} & \textbf{29.34} & \textbf{4.64} & \textbf{0.189} \\
\cmidrule(lr){1-10}
\multirow{4}{*}{\shortstack{LLaVA\\with CXR-CLIP}} 
& Image-Only & 0.213 & 0.174 & 0.101 & 0.123 & 0.151 & 19.20 & 4.69 & 0.185 \\
& + Concepts & 0.328 & 0.330 & 0.161 & 0.230 & 0.172 & 19.62 & 4.87 & 0.192 \\
& + RAG & 0.489 & 0.518 & 0.309 & 0.442 & 0.181 & 25.53 & 5.58 & 0.198 \\
& + \ModelName & \textbf{0.498} & \textbf{0.526} & \textbf{0.314} & \textbf{0.443} & \textbf{0.187} & \textbf{29.78} & \textbf{6.08} & \textbf{0.201} \\
\midrule
\multicolumn{10}{c}{\textit{Supervised Fine-Tuning}} \\
\midrule
\multirow{4}{*}{LLaVA-Med} 
& Image-Only  & 0.470 & 0.486 & 0.266 & 0.392 & 0.174 & 27.81 & 7.43 & 0.211 \\
& + Concepts  & 0.476 & 0.502 & 0.284 & 0.412 & \textbf{0.188} & 28.83 & \textbf{7.87} & \textbf{0.220} \\
& + RAG  & 0.477 & 0.499 & 0.287 & 0.412 & 0.176 & 30.50 & 7.26 & 0.212 \\
& + \ModelName  & \textbf{0.488} & \textbf{0.510} & \textbf{0.301} & \textbf{0.424} & 0.177 & \textbf{30.81} & 7.47 & 0.213 \\
\cmidrule(lr){1-10}
\multirow{4}{*}{\shortstack{LLaVA\\with CXR-CLIP}} 
& Image-Only  & 0.393 & 0.448 & 0.215 & 0.343 & 0.161 & 22.38 & 5.60 & 0.207 \\
& + Concepts  & 0.486 & 0.510 & 0.283 & 0.413 & \textbf{0.184} & 28.17 & \textbf{7.57} & \textbf{0.225} \\
& + RAG  & 0.477 & 0.501 & 0.293 & 0.414 & 0.168 & 30.16 & 6.84 & 0.204 \\
& + \ModelName & \textbf{0.488} & \textbf{0.512} & \textbf{0.300} & \textbf{0.423} & 0.180 & \textbf{30.49} & 6.98 & 0.206 \\
\bottomrule
\end{tabular}
}
\end{table*}

\paragraph*{Zero-Shot.} 
In the \emph{Zero-Shot} setting, the LLaVA-Med baseline exhibits clear limitations for RRG. 
Despite leveraging general medical pretraining, it attains an F1-Radgraph (F1-RG) of only $0.052$, indicating poor alignment of clinical entities and relations, and very low NLP metrics, reflecting limited lexical overlap with reference reports. 
CheXbert Micro-F1$_{14}$ reaches $0.255$, suggesting some ability to recognize common pathologies but insufficient overall clinical reliability.
Augmenting the prompt with concepts or retrieval leads to a consistent progression in performance.
Adding concepts improves both clinical and textual metrics: F1-RG rises to $0.073$, CheXbert Micro-F1$_{14}$ to $0.401$, and ROUGE-L to $0.146$, indicating that even in a \emph{Zero-Shot} regime, visual concepts provide useful structured cues about relevant findings.
RAG yields substantially larger gains, with F1-RG increasing to $0.184$ and CheXbert Micro-F1$_{14}$ to $0.498$.
These gains demonstrate that retrieved similar cases provide rich contextual and linguistic information that enhances both clinical accuracy and report fluency.
\ModelName achieves the best overall \emph{Zero-Shot} performance for LLaVA-Med, slightly but consistently outperforming RAG across all clinical and lexical metrics, indicating that integrating concepts with retrieved reports yields an additional performance benefit.

A similar trend is observed for the configuration using CXR-CLIP as the visual encoder, which starts from a stronger \emph{Zero-Shot} baseline in terms of fluency as a result of the one-epoch alignment of its projection layer on MIMIC-CXR.
The baseline achieves an F1-RG of $0.151$ and BLEU-4 of $4.69$, considerably higher than LLaVA-Med, whilst CheXbert Micro-F1$_{14}$ remains lower at $0.213$. 
This pattern suggests that the alignment primarily improves generic report structure and phrasing, but does not immediately translate into superior pathology coverage. 
As in the previous case, both concept and retrieval augmentation are beneficial: \emph{Concepts} strategy increases CheXbert Micro-F1$_{14}$ to $0.328$ and F1-RG to $0.172$, whereas RAG raises these values to $0.489$ and $0.181$, respectively. 
\ModelName configuration attains again the best overall \emph{Zero-Shot} performance in this setting, with F1-RG of $0.187$, CheXbert Micro-F1$_{14}$ of $0.498$, and the highest BLEU-1 ($29.78$), BLEU-4 ($6.08$), and ROUGE-L ($0.201$).
Overall, these \emph{Zero-Shot} results indicate that concept-based and retrieval-based augmentation contribute positively across both architectures, and that their combination systematically improves clinical and lexical metrics over Image-Only and single-augmentation baselines.

\paragraph*{SFT.}
Under \emph{SFT}, the \emph{Image-Only} baselines of both architectures exhibit substantial improvements over their \emph{Zero-Shot} counterparts, confirming the importance of task-specific adaptation for RRG.
The LLaVA-Med baseline shows substantial gains across all metrics, with F1-RG increasing from $0.052$ to $0.174$, CheXbert Micro-F1$_{14}$ from $0.255$ to $0.470$, and BLEU-4 from $0.71$ to $7.43$, with ROUGE-L reaching $0.211$. 
On top of this stronger baseline, the three augmentation strategies remain beneficial, but their role shifts compared to the \emph{Zero-Shot} setting.
Adding only concepts yields the highest F1-RG in this configuration ($0.188$) and the best BLEU-4 and ROUGE-L scores ($7.87$ and $0.220$), along with moderate gains in CheXbert Micro- and Macro-F1 ($0.476$ and $0.284$). 
In this setting, \emph{Concepts} strategy yields gains on CheXbert while having a larger impact on sequence-level metrics such as F1-RG, BLEU-4, and ROUGE-L, suggesting that explicit visual concepts are especially beneficial for structuring clinically detailed reports.
RAG shows a different pattern: F1-RG remains close to the baseline ($0.176$ vs. $0.174$), whereas BLEU-1 increases to $30.50$. 
This pattern suggests that, once the model has been adapted on MIMIC-CXR using retrieval-augmented prompts, the additional in-domain retrieved reports mainly enrich lexical diversity and increase report length, without yielding proportional gains on metrics that are more sensitive to clinical structure and relational correctness.
Finally, the combined \ModelName achieves the strongest overall performance: it attains the highest CheXbert Micro-F1$_{14}$ of $0.488$ and Macro-F1$_{14}$ of $0.301$, along with the highest BLEU-1 score of $30.81$. 
These results confirm that interpretability-driven augmentation provides complementary benefits to \emph{SFT}, specifically for clinical accuracy metrics that directly measure factual correctness in generated reports.

For the CXR-CLIP configuration, the \emph{SFT} baseline remains weaker than the corresponding LLaVA-Med baseline on MIMIC-CXR (e.g., CheXbert Micro-F1$_{14}$ of $0.393$ vs. $0.470$), reflecting the advantage of LLaVA-Med’s medically pretrained language component.
The augmentation strategies follow the same progressive improvement pattern observed in other configurations. 
\emph{Concepts} augmentation produces dramatic gains, with F1-RG jumping to $0.184$ and CheXbert Micro-F1$_{14}$ reaching $0.488$, substantially exceeding the improvements observed with LLaVA-Med and nearly recovering the baseline performance deficit. 
Notably, this configuration achieves the highest BLEU-4 score of $7.57$ across all \emph{SFT} CXR-CLIP settings and the highest overall ROUGE-L of $0.225$, mirroring the pattern observed with LLaVA-Med.
These results further demonstrate that interpretable visual concepts offer structured information that effectively complements visual encoder features.
RAG-only yields mixed results, with F1-RG at $0.168$ and CheXbert Macro-F1$_{14}$ reaching $0.293$, while CheXbert Micro-F1$_{14}$ of $0.477$ remains below the \emph{Concepts} configuration.
The combined \ModelName approach achieves balanced performance with the highest CheXbert Micro-F1$_{14}$ at $0.488$ and Macro-F1$_{14}$ at $0.300$, alongside the highest BLEU-1 of $30.49$.

\paragraph*{MIMIC-CXR Summary.}
Taken together, the MIMIC-CXR experiments reveal two main patterns regarding the interaction between augmentation strategies and training regimes.

First, the effectiveness of RAG is strongly regime-dependent.
In the \emph{Zero-Shot} setting, adding in-domain retrieved reports substantially improves clinical and linguistic metrics, consistent with models that rely heavily on external context to compensate for limited task-specific adaptation. 
In the \emph{SFT} regime, where models are trained with retrieved reports in the prompt, the role of retrieval changes. 
While unigram coverage (BLEU-1) improves substantially, metrics sensitive to longer-range structure (BLEU-4, ROUGE-L) underperform relative to \emph{Concepts} augmentation.
Furthermore, clinical metrics exhibit modest yet consistent degradation compared to \emph{Zero-Shot} for both RAG and \ModelName.
This behaviour suggests that when supervision and retrieval are drawn from the same in-domain distribution, retrieval provides diminishing returns as the model internalizes distributional patterns present in the retrieved context.

Secondly, SpLiCE-derived concepts contribute consistently across both regimes. 
In \emph{Zero-Shot} configurations, \emph{Concepts} alone clearly improve pathology identification and F1-RG. 
In the \emph{SFT} setting, SpLiCE contributes more evidently on metrics that capture the quality of clinically complex sequences, while the gain on CheXbert F1 is comparatively smaller. 

The combined approach, \ModelName, which combines concept and retrieval augmentation, leverages the complementary strengths of both: in \emph{Zero-Shot}, it slightly amplifies the benefits of RAG; in the \emph{SFT} regime, it maintains or improves aggregate clinical performance while attenuating some of the redundancy associated with retrieval alone.

Finally, the systematic gap between CheXbert Micro-F1 and Macro-F1 across all methods reflects the pronounced class imbalance in chest X-ray reports. 
The improvements in Macro-F1 observed with \ModelName indicate that the method contributes to a better coverage of underrepresented findings, supporting its suitability for clinically realistic reporting scenarios.

\subsubsection{IU X-ray}\label{IUxrayResults}
\cref{tab:iuxray_results} presents quantitative results on the IU X-ray dataset. 
This dataset, differently from MIMIC-CXR, features considerably shorter and more concise reports, and employs a more limited medical vocabulary.
In this case, our experimental setup implements cross-domain retrieval, where similar cases are retrieved from the MIMIC-CXR database rather than from IU X-ray's own training set. 
This configuration puts our framework's ability to generalize across different datasets and reporting styles to the test, since the retrieved context originates from a different institutional source with different imaging protocols and documentation practices.

\begin{table*}[h]
\centering
\caption{Quantitative results on the IU X-Ray test set for two model configurations
(LLaVA-Med and LLaVA with CXR-CLIP) and two training regimes
(\textit{Zero-Shot} and \textit{Supervised Fine-Tuning}).
We report F1-RadGraph (F1-RG), and NLP metrics (BLEU-1 as B-1, BLEU-4 as B-4, ROUGE-L as R-L). CheXbert-based label metrics are reported over 14 labels (Micro-F1$_{14}$, Macro-F1$_{14}$) and over the 5 most prevalent findings (Micro-F1$_{5}$, Macro-F1$_{5}$).
Retrieval is performed cross-domain from the MIMIC-CXR training set.
}
\label{tab:iuxray_results}
\resizebox{\textwidth}{!}{
\begin{tabular}{llcccccccc}
\toprule
\textbf{Model} & \textbf{Method} & \textbf{Micro-F1}$_{14}$ & \textbf{Micro-F1}$_{5}$ & \textbf{Macro-F1}$_{14}$ & \textbf{Macro-F1}$_{5}$ & \textbf{F1-RG} & \textbf{B-1} & \textbf{B-4} & \textbf{R-L} \\
\midrule
\multicolumn{10}{c}{\textit{Zero-shot}} \\
\midrule
\multirow{4}{*}{LLaVA-Med} 
& Image-Only & 0.063 & 0.042 & 0.047 & 0.019 & 0.074 & 17.36 & 1.08 & 0.125 \\
& + Concepts & 0.122 & 0.162 & 0.092 & 0.134 & 0.064 & 13.79 & 0.85 & 0.111 \\
& + RAG & 0.377 & 0.344 & 0.220 & 0.271 & 0.228 & 21.04 & 3.60 & 0.177 \\
& + \ModelName & \textbf{0.387} & \textbf{0.397} & \textbf{0.252} & \textbf{0.315} & \textbf{0.234} & \textbf{24.34} & \textbf{4.10} & \textbf{0.191} \\
\cmidrule(lr){1-10}
\multirow{4}{*}{\shortstack{LLaVA\\with CXR-CLIP}} 
& Image-Only & 0.305 & 0.082 & 0.038 & 0.043 & 0.188 & 5.56 & 1.26 & 0.168 \\
& + Concepts & 0.307 & 0.182 & 0.085 & 0.087 & 0.199 & 8.15 & 1.98 & 0.199 \\
& + RAG & 0.367 & 0.343 & 0.212 & 0.247 & 0.213 & 25.35 & 6.13 & 0.203 \\
& + \ModelName & \textbf{0.378} & \textbf{0.361} & \textbf{0.232} & \textbf{0.298} & \textbf{0.247} & \textbf{27.84} & \textbf{6.75} & \textbf{0.221} \\
\midrule
\multicolumn{10}{c}{\textit{Supervised Fine-Tuning}} \\
\midrule
\multirow{4}{*}{LLaVA-Med} 
& Image-Only & 0.326 & 0.115 & 0.031 & 0.059 & 0.175 & 14.25 & 4.50 & 0.179 \\
& + Concepts & 0.336 & 0.185 & 0.082 & 0.081 & 0.178 & 23.21 & 5.67 & 0.181 \\
& + RAG & 0.468 & 0.356 & 0.183 & 0.205 & 0.249 & 28.15 & 7.73 & 0.251 \\
& + \ModelName & \textbf{0.501} & \textbf{0.526} & \textbf{0.244} & \textbf{0.355} & \textbf{0.252} & \textbf{28.37} & \textbf{8.00} & \textbf{0.252} \\
\cmidrule(lr){1-10}
\multirow{4}{*}{\shortstack{LLaVA\\with CXR-CLIP}} 
& Image-Only & 0.376 & 0.102 & 0.037 & 0.053 & 0.235 & 14.25 & 4.50 & 0.235 \\
& + Concepts & 0.427 & 0.362 & 0.118 & 0.178 & 0.244 & 18.83 & 6.06 & 0.242 \\
& + RAG & 0.468 & 0.395 & 0.172 & 0.245 & 0.244 & 22.58 & 6.91 & 0.243 \\
& + \ModelName & \textbf{0.486} & \textbf{0.439} & \textbf{0.174} & \textbf{0.256} & \textbf{0.248} & \textbf{22.90} & \textbf{7.10} & \textbf{0.249} \\
\bottomrule
\end{tabular}
}
\end{table*}

\paragraph*{Zero-Shot.}
In the \emph{Zero-Shot} setting, the LLaVA-Med baseline performs very poorly on IU X-Ray, with a CheXbert Micro-F1$_{14}$ of $0.063$ and an F1-RG of $0.074$, confirming that general medical pretraining does not directly translate into effective RRG for this dataset.
The \emph{Concepts} condition produces a mixed effect: CheXbert Micro-F1$_{14}$ roughly doubles to $0.122$, indicating improved detection of some pathologies, yet F1-RG decreases slightly to $0.064$ and all NLP metrics deteriorate (e.g., BLEU-1 drops from $17.36$ to $13.79$).
This behaviour is consistent with a style mismatch: when concepts are injected without further constraints, LLaVA-Med tends to expand each keyword into lengthy explanatory sentences. This contrasts with the highly concise IU X-ray references, reducing n-gram overlap despite potentially correct clinical content.
In contrast, cross-domain retrieval yields substantial improvements. RAG on MIMIC-CXR increases CheXbert Micro-F1$_{14}$ to $0.377$, F1-RG to $0.228$, and BLEU-4 to $3.60$, indicating that retrieved examples provide useful templates despite originating from a different institution and reporting style.
Finally, \ModelName achieves the best overall \emph{Zero-Shot} performance for LLaVA-Med (Micro-F1$_{14}$ $0.387$, F1-RG $0.234$, BLEU-4 $4.10$), suggesting that concepts help the model focus on clinically salient parts of the retrieved reports and partially counteract the verbosity that arises when concepts are used in isolation.

For LLaVA with CXR-CLIP configuration, the \emph{Zero-Shot} baseline attains CheXbert Micro-F1$_{14}$ = $0.305$ and F1-RG = $0.188$, substantially higher than LLaVA-Med. This reflects the one-epoch projector alignment on IU X-ray, which provides better dataset-specific adaptation compared to LLaVA-Med's general medical pretraining. On the other hand, performance on the 5-label subset is extremely weak (Micro-F1$_5$ = 0.082) and several Macro-F1 scores are low, indicating uneven pathology recognition across label subsets. NLP metrics are also poor (BLEU-1 = $5.56$, BLEU-4 = $1.26$), showing that this short alignment phase only partially adapts the model to the concise IU X-ray style.
\emph{Concepts} augmentation provides modest yet consistent gains (F1-RG = $0.199$, slight improvements in CheXbert metrics), though effectiveness is limited by the projector alignment on IU X-ray's small training set.
RAG produces much larger benefits: Micro-F1$_{14}$ rises to $0.367$, F1-RG to $0.213$, and BLEU-1 jumps to $25.35$, confirming that cross-domain retrieval from MIMIC-CXR supplies useful clinical templates and linguistic structure. 
The combined \ModelName condition yields the strongest \emph{Zero-Shot} performance for this configuration, with Micro-F1$_{14}$ = $0.378$, F1-RG = $0.247$, BLEU-1 = $27.84$, BLEU-4 = $6.75$, and ROUGE-L = $0.221$. This indicates that concept-level signals derived from SpLiCE help the model exploit cross-domain retrieved context more selectively, improving both clinical accuracy and report quality.

\paragraph*{SFT.}
In the \emph{SFT} setting, LLaVA-Med exhibits substantial gains over its \emph{Zero-Shot} performance on IU X-Ray.
The baseline Micro-F1$_{14}$ increases from 0.063 to 0.326 and F1-RG from $0.074$ to $0.175$, confirming that even a relatively small amount of supervision is sufficient to substantially improve task-specific behaviour on a new dataset. 
On top of this, the augmentation strategies continue to provide consistent benefits. \emph{Concepts} augmentation yields modest improvements, whereas RAG leads to more pronounced gains, reaching $0.468$ Micro-F1$_{14}$ and $0.249$ F1-RG. 
The combined \ModelName condition achieves the strongest overall performance, with $0.501$ Micro-F1$_{14}$, $0.252$ F1-RG, and BLEU-4 of $8.00$. 
Unlike the in-domain setting on MIMIC-CXR, where RAG becomes partially redundant after \emph{SFT}, here cross-domain retrieval from MIMIC-CXR continues to supply complementary information that is not fully captured by supervised training on the much smaller IU X-Ray corpus, and \ModelName is able to exploit this additional signal more effectively.

For LLaVA with CXR-CLIP configuration, \emph{SFT} on IU X-ray leads to analogous trends. 
The \emph{SFT} baseline reaches CheXbert Micro-F1$_{14}$ = $0.376$ and F1-RG = $0.235$;
\emph{Concepts} augmentation improves sequence-level metrics, with F1-RG = $0.244$, BLEU-4 = $6.06$, and ROUGE-L = $0.242$;
RAG further increases label-based scores (Micro-F1$_{14}$ = $0.468$, F1-RG = $0.244$).
The full \ModelName configuration yields a balanced improvement, with Micro-F1$_{14}$ = $0.486$, F1-RG = $0.248$, and BLEU-4 = $7.10$, combining the benefits of concept guidance and retrieval-based context.
Despite the comparable clinical metrics under \ModelName, a clear discrepancy persists in lexical quality: LLaVA-Med attains substantially higher BLEU-1 ($28.37$ vs. $22.90$). This is plausibly due to its medically pretrained language component, which starts from a richer medical vocabulary and a broader set of reporting patterns. In the CXR-CLIP variant, the Mistral-7B backbone is adapted only on the small IU X-ray corpus, limiting its ability to acquire diverse and specialised radiology phrasing. As a result, report generations remain less varied and exhibit lower n-gram overlap, even when clinical content is comparable.

\paragraph*{IU X-ray Summary.}
Overall, the IU X-ray experiments show that the proposed augmentation strategies remain effective in a low-resource, cross-domain scenario. \emph{SFT} substantially improves both backbones, cross-domain RAG continues to offer clear benefits rather than becoming redundant, and \ModelName consistently matches or exceeds the single augmentations. At the same time, the persistent gap in BLEU-1 between LLaVA-Med and CXR-CLIP configurations underscores the practical value of a medically pretrained language module for producing lexically rich reports when task-specific supervision is scarce.

\subsection{Qualitative Results}\label{QualResults}
While quantitative metrics capture global trends across datasets and configurations, qualitative analysis helps clarify how concept extraction and retrieval affect individual predictions. In this section, we analyse two representative chest X-ray cases from MIMIC-CXR using the LLaVA-Med configuration. 
The first illustrates typical \emph{Zero-Shot} failure modes and their mitigation with \ModelName, while the second examines how the same mechanisms behave after \emph{SFT}.
For each case, we compare reports produced by the Image-Only, \emph{Concepts}, RAG, and \ModelName conditions, and highlight patterns of hallucination, omission, superfluity, as well as clinically accurate description.

\subsubsection{Generated Report Comparison}

\paragraph*{Zero-Shot.}
\cref{tab:prompt_case} shows a \emph{Zero-Shot} example on MIMIC-CXR that reflects the error patterns observed in \cref{tab:mimic_results}. 
The ground truth report describes multiple devices (endotracheal tube, orogastric tube, right internal jugular catheter), low lung volumes, and a right upper lobe opacity suspicious for pneumonia, explicitly ruling out pleural effusion and pneumothorax.

\begin{table*}[h]
\centering
\caption{Qualitative Comparison in \emph{Zero-Shot} Setting for LLaVA-Med. Highlighting indicates: \hlincorrect{incorrect/hallucinated findings}, \hlincomplete{incomplete descriptions}, \hlsuperfluous{superfluous details}, \hlaccurate{accurate content}.}
\label{tab:prompt_case}
\small
\resizebox{0.8\textwidth}{!}{
\begin{tabular}{m{3.5cm}m{11cm}}
\toprule
\multicolumn{2}{c}{\textsc{Zero-Shot Setting}} \\
\toprule
\textbf{Chest X-Ray Image} & \textbf{Ground Truth Report} \\
\midrule
\raisebox{-0.9\height}{\includegraphics[width=3cm]{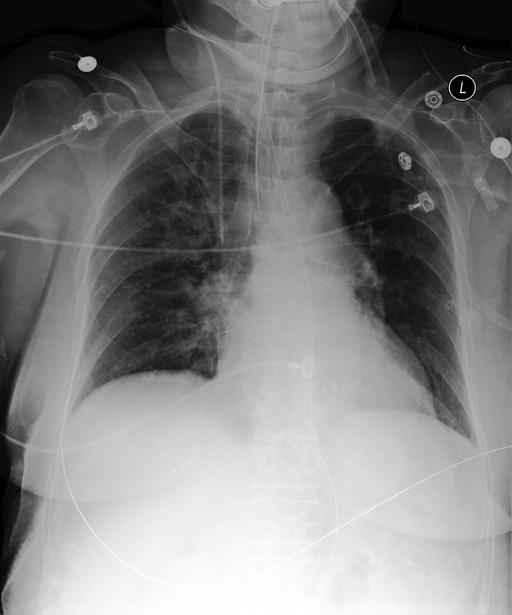}} & 
\parbox[t]{11cm}{Endotracheal tube terminates 2 cm above the carina. Orogastric tube terminates in the stomach. Right internal jugular catheter terminates in the mid SVC. Lungs are low in volume with stable right upper lung opacities which are better assessed on the recent chest CT but suspicious for pneumonia. There is no pneumothorax or pleural effusion. Heart is normal in size. Normal cardiomediastinal silhouette.} \\
\cmidrule(lr){2-2}
& \textbf{SpLiCE Concepts:} \textit{pulmonary vascularity, overt pulmonary, right upper, endotracheal tube, internal jugular} \\
\midrule
\textbf{Image-Only} & 
The chest X-ray image shows \hlincorrect{bilateral diffuse infiltrates, which are abnormal substances or collections of cells that have entered the lung tissue}. These infiltrates can be seen in both lungs and are distributed throughout the lung fields. \\
\midrule
\textbf{+ Concepts} & 
The chest X-ray shows \hlincorrect{overt pulmonary edema}, which is a condition where there is an abnormal accumulation of fluid in the lungs. Additionally, there is a \hlaccurate{right upper lobe opacity}, which could be due to various reasons such as infection, inflammation, or other lung abnormalities. The internal jugular vein is visible, and an \hlaccurate{endotracheal tube is in place}, which is a \hlsuperfluous{tube that is inserted through the mouth or nose and into the trachea to help maintain an open airway and assist with breathing}. Lastly, the pulmonary vascularity appears to be normal. \\
\midrule
\textbf{+ RAG} & 
\hlincorrect{The chest X-ray findings are similar to those in the earlier study}. The main differences observed are the placement of a \hlincorrect{left IJ catheter} and the presence of an \hlaccurate{indistinct area of opacification in the right upper zone medially}, which could represent a developing consolidation. Additionally, there is a \hlaccurate{right IJ central venous catheter} with its tip at the lower SVC near the cavoatrial junction, and the \hlaccurate{endotracheal tube} has been slightly advanced. \hlaccurate{No pneumothorax} is noted. \hlincorrect{Pulmonary vascular congestion is observed}. \\
\midrule
\textbf{+ \ModelName} & 
In the \hlaccurate{right upper zone, there is an indistinct area of opacification}, which could represent a developing consolidation. Additionally, there is a \hlaccurate{right internal jugular central venous catheter} with its tip at the lower superior vena cava near the cavoatrial junction. The \hlaccurate{endotracheal tube} has been slightly advanced, with its tip 2.2 cm above the carina. \hlaccurate{No pneumothorax} is noted, and the \hlincorrect{pulmonary vascular congestion is present}. \hlaccurate{The cardiomediastinal silhouette remains unchanged.} \\
\bottomrule
\end{tabular}
}
\end{table*}

The baseline LLaVA-Med model produces a markedly incorrect description, hallucinating bilateral diffuse infiltrates and failing to mention any of the indwelling devices, in line with the very low \emph{Zero-Shot} F1-RG observed quantitatively. This behaviour indicates that, without additional guidance, the model tends to generate generic patterns of abnormality that are not well grounded in the specific image.
With \emph{Concepts} augmentation, all extracted keywords (e.g., ``endotracheal tube'', ``internal jugular'', ``right upper'') are explicitly mentioned in the generated text, and the report correctly identifies both the endotracheal tube and a right upper lobe opacity. However, the model now over-interprets the concept set by asserting overt pulmonary edema, a finding not present in the reference report. This illustrates how SpLiCE decomposition can substantially increase CheXbert Micro-F1 by encouraging the mention of clinically salient terms, while still lagging behind retrieval-based strategies on metrics that reward correct entity–relation structure. Notably, the model produces verbose explanations of medical concepts (e.g., describing what an endotracheal tube does), as LLaVA-Med's general medical pretraining has not been adapted to the concise technical language conventions of radiology reports.
RAG-only reduces some of these concept-driven hallucinations but introduces others. Conditioning on retrieved in-domain reports leads the model to blend details from similar but non-identical cases, such as mentioning a left internal jugular catheter and pulmonary vascular congestion that are not supported by the target image. This exemplifies a typical retrieval-induced failure mode in which information from nearest neighbours leaks into the generated report.
The combined \ModelName condition achieves the most clinically plausible description. It correctly identifies the right internal jugular catheter, the endotracheal tube position, and the right upper zone opacity, and avoids the more severe hallucinations observed in the RAG-only report, although it still mentions mild vascular congestion. 
This example mirrors the quantitative trends: concept cues alone can act as over-strong priors, and retrieval alone can import spurious details, whereas their combination yields better grounded descriptions that more closely match the reference report.

\paragraph*{SFT.}
The second example (\cref{tab:sft_case}) illustrates behaviour in the \emph{SFT} setting on MIMIC-CXR. 
The reference report describes low lung volumes and bibasilar opacities most likely representing atelectasis, and explicitly states the absence of pleural effusion or pneumothorax, with a stable and unremarkable cardiomediastinal silhouette.

\begin{table*}[h]
\centering
\caption{Qualitative Comparison in \emph{Supervised Fine-Tuning} Setting or LLaVA-Med. Highlighting indicates: \hlincorrect{incorrect/hallucinated findings}, \hlincomplete{incomplete descriptions}, \hlsuperfluous{superfluous details}, \hlaccurate{accurate content}.}
\label{tab:sft_case}
\small
\resizebox{0.8\textwidth}{!}{
\begin{tabular}{m{3.5cm}m{11cm}}
\toprule
\multicolumn{2}{c}{\textsc{Supervised Fine-Tuned Setting}} \\
\toprule
\textbf{Chest X-Ray Image} & \textbf{Ground Truth Report} \\
\midrule
\raisebox{-0.9\height}{\includegraphics[width=3cm]{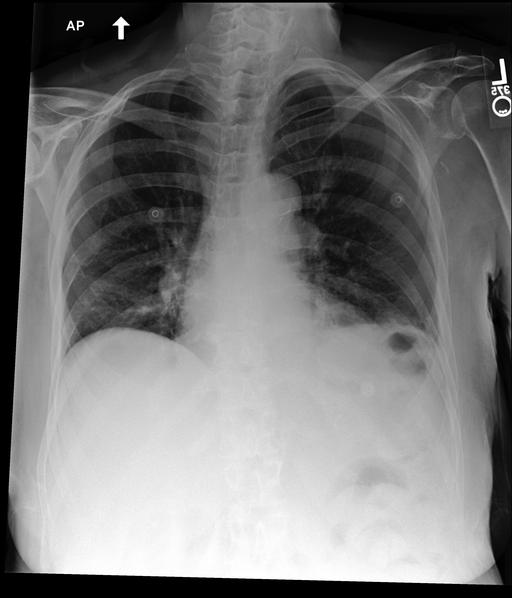}} & 
\parbox[t]{11cm}{Frontal and lateral views of the chest were obtained. There are low lung volumes which accentuate the bronchovascular markings. Bibasilar opacities are seen, which most likely represent atelectasis, although aspiration or infection are not excluded in the appropriate clinical setting. No pleural effusion or pneumothorax is seen. The cardiac and mediastinal silhouettes are stable and unremarkable.} \\
\cmidrule(lr){2-2}
& \textbf{SpLiCE Concepts:} \textit{lateral view, lung base, volume low, left hemidiaphragm, appropriate clinical} \\
\midrule
\textbf{Image-Only} & 
The heart size is normal. The hilar and mediastinal contours are normal. There is a \hlincorrect{new} \hlincomplete{left lower lobe opacity, concerning for pneumonia}. No pleural effusion or pneumothorax. \hlincomplete{The visualized osseous structures are unremarkable.} \\
\midrule
\textbf{+ Concepts} & 
In comparison with the study of \_\_\_, there are \hlaccurate{lower lung volumes}. Cardiac silhouette is within normal limits. \hlincomplete{Opacification at the left base is consistent with atelectatic changes} and \hlincorrect{possible small effusion}. \hlaccurate{In the appropriate clinical setting, superimposed pneumonia would have to be considered}. \\
\midrule
\textbf{+ RAG} & 
\hlincorrect{The lungs are clear without focal consolidation}. \hlaccurate{No pleural effusion or pneumothorax is seen}. \hlaccurate{The cardiac and mediastinal silhouettes are unremarkable}. \\
\midrule
\textbf{+ \ModelName} & 
\hlaccurate{Lung volumes are low}. \hlaccurate{Mediastinal and hilar contours are unremarkable}. Pulmonary vasculature is not engorged. \hlaccurate{Patchy opacities in the lung bases likely reflect areas of atelectasis}. \hlaccurate{No pleural effusion or pneumothorax is present}. No acute osseous abnormality is visualized. \\
\bottomrule
\end{tabular}
}
\end{table*}

The Image-Only baseline captures part of this picture: it correctly rules out pleural effusion and pneumothorax and describes normal cardiac and mediastinal contours, but introduces a clinically relevant error by emphasising a \textit{new} left lower lobe opacity concerning for pneumonia and omitting any mention of low lung volumes. This combination of partial correctness and misplaced emphasis is consistent with the moderate CheXbert Micro-F1 and F1-RG achieved by the \emph{SFT} baseline in \cref{tab:mimic_results}.
With \emph{Concepts} augmentation, the model more faithfully reflects the ground truth: it explicitly recognises lower lung volumes and describes left basal opacification as atelectatic change, while conditionally mentioning possible pneumonia ``in the appropriate clinical setting'', closely echoing the phrasing of the reference report. The main discrepancy is the mention of a possible small effusion, which is explicitly ruled out in the ground truth. Overall, this aligns with the observed improvements in F1-RG and sequence-level metrics under SpLiCE.
In the RAG-only condition, the report fails to mention the bibasilar opacities and instead states that the lungs are clear without focal consolidation. This represents a clinically concerning omission and is consistent with a failure mode in which the model appears to rely too heavily on retrieved studies and does not fully ground its description in the current image.
The combined \ModelName report strikes a better balance: it correctly identifies low lung volumes and patchy basal opacities likely reflecting atelectasis, preserves the correct absence of pleural effusion and pneumothorax, and maintains a coherent description of the cardiomediastinal silhouette. 
This progression from baseline to \ModelName mirrors the quantitative gains reported for MIMIC-CXR and qualitatively illustrates how concept-level signals can prevent complete omission of pathologies when retrieval alone is unreliable.

\subsubsection{Interpretability Through Concept Visualization}

Beyond improvements in clinical accuracy, a central motivation of \ModelName is to make the visual evidence underlying generated reports explicitly inspectable. In this section, we show how SpLiCE concepts and gradient-based explanations can be combined to reveal where the model “looks” in the image when producing specific medical terms in the report.

\begin{table*}[h]
\centering
\caption{Qualitative Examples with Concept-Specific Visual Explanations. Grad-ECLIP heat maps highlight image regions that influence the LLaVA-Med encoder's matching between input images and SpLiCE concepts (highlighted in blue). The visualizations show which spatial features contribute to concept identification in the generated reports. Top row: Zero-Shot setting; bottom row: supervised fine-tuning setting.}
\label{tab:gradeclip_examples}
\footnotesize
\resizebox{0.8\textwidth}{!}{
\setlength{\tabcolsep}{6pt}
\renewcommand{\arraystretch}{1.2}
\begin{tabular}{|p{6cm}|p{6cm}|}
\hline
\rowcolor{blue!10}
\textbf{Ground Truth} & \textbf{Generated Report} \\
\hline
\underline{Endotracheal tube} terminates 2 cm above the carina. Orogastric tube terminates in the stomach. Right internal jugular catheter terminates in the mid SVC. Lungs are low in volume with stable \underline{right upper} lung opacities which are better assessed on the recent chest CT but suspicious for pneumonia. There is no pneumothorax or pleural effusion. Heart is normal in size. Normal cardiomediastinal silhouette. 
&
The chest X-ray shows several findings. In the \colorbox{blue!20}{\textcolor{blue!90}{\textbf{right upper}}} zone, there is an indistinct area of opacification, which could represent a developing consolidation. Additionally, there is a right internal jugular central venous catheter with its tip at the lower superior vena cava near the cavoatrial junction. The \colorbox{blue!20}{\textcolor{blue!90}{\textbf{endotracheal tube}}} has been slightly advanced, with its tip 2.2 cm above the carina. No pneumothorax is noted, and the pulmonary vascular congestion is present. The cardiomediastinal silhouette remains unchanged. \\
\hline
\multicolumn{2}{|c|}{
\cellcolor{blue!5}
\begin{tabular}{c c@{\hspace{0.6cm}}c}
\textbf{Original Image} & \multicolumn{2}{c}{\textbf{Grad-ECLIP Activation Maps}} \\
\rule{0pt}{0.5cm}
\fbox{\includegraphics[width=3.5cm]{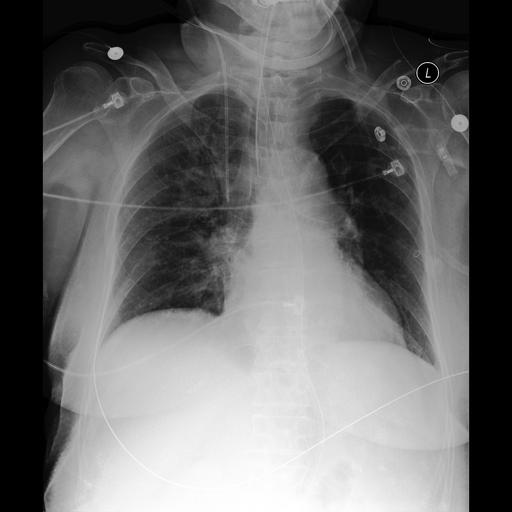}} &
\fbox{\includegraphics[width=3.5cm]{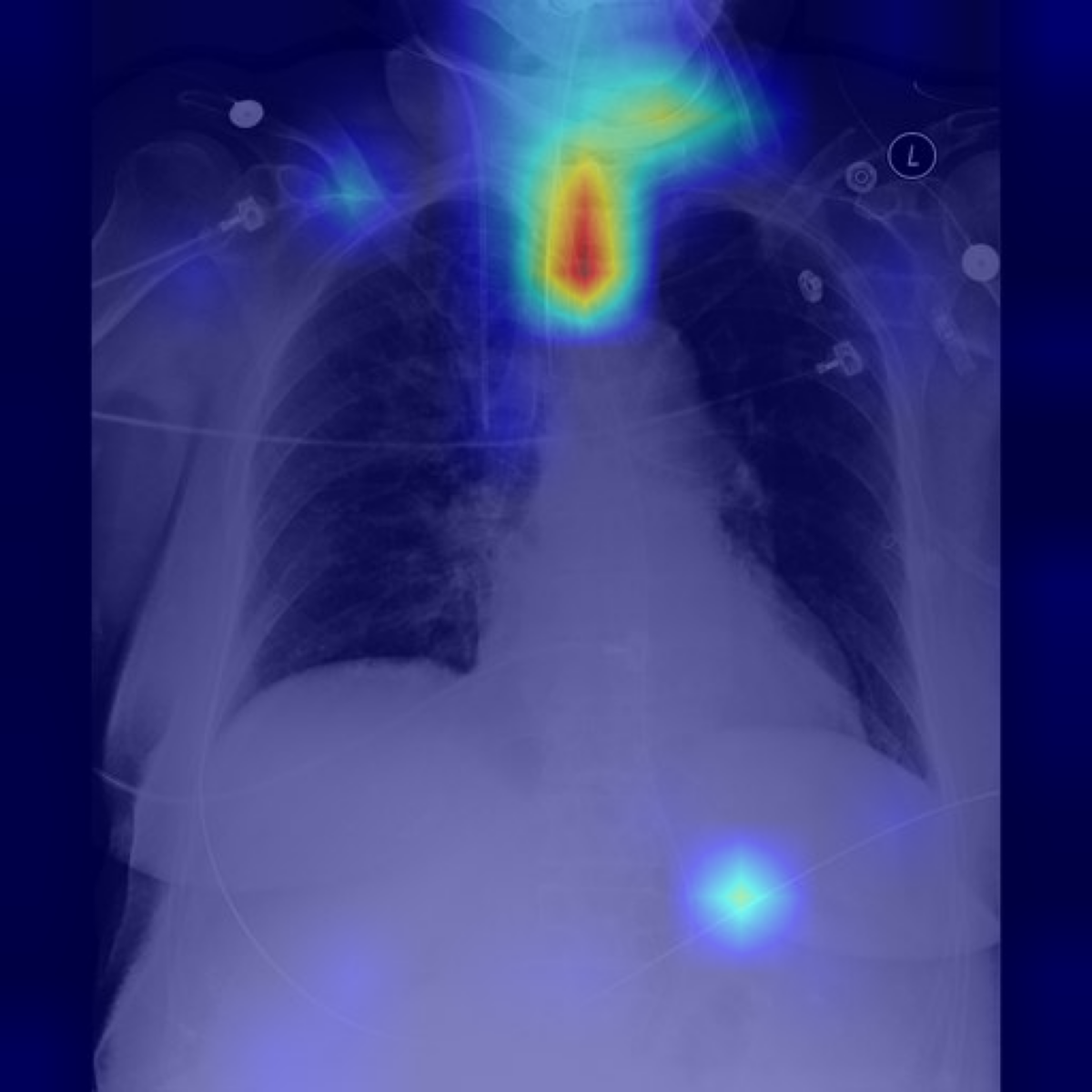}} &
\fbox{\includegraphics[width=3.5cm]{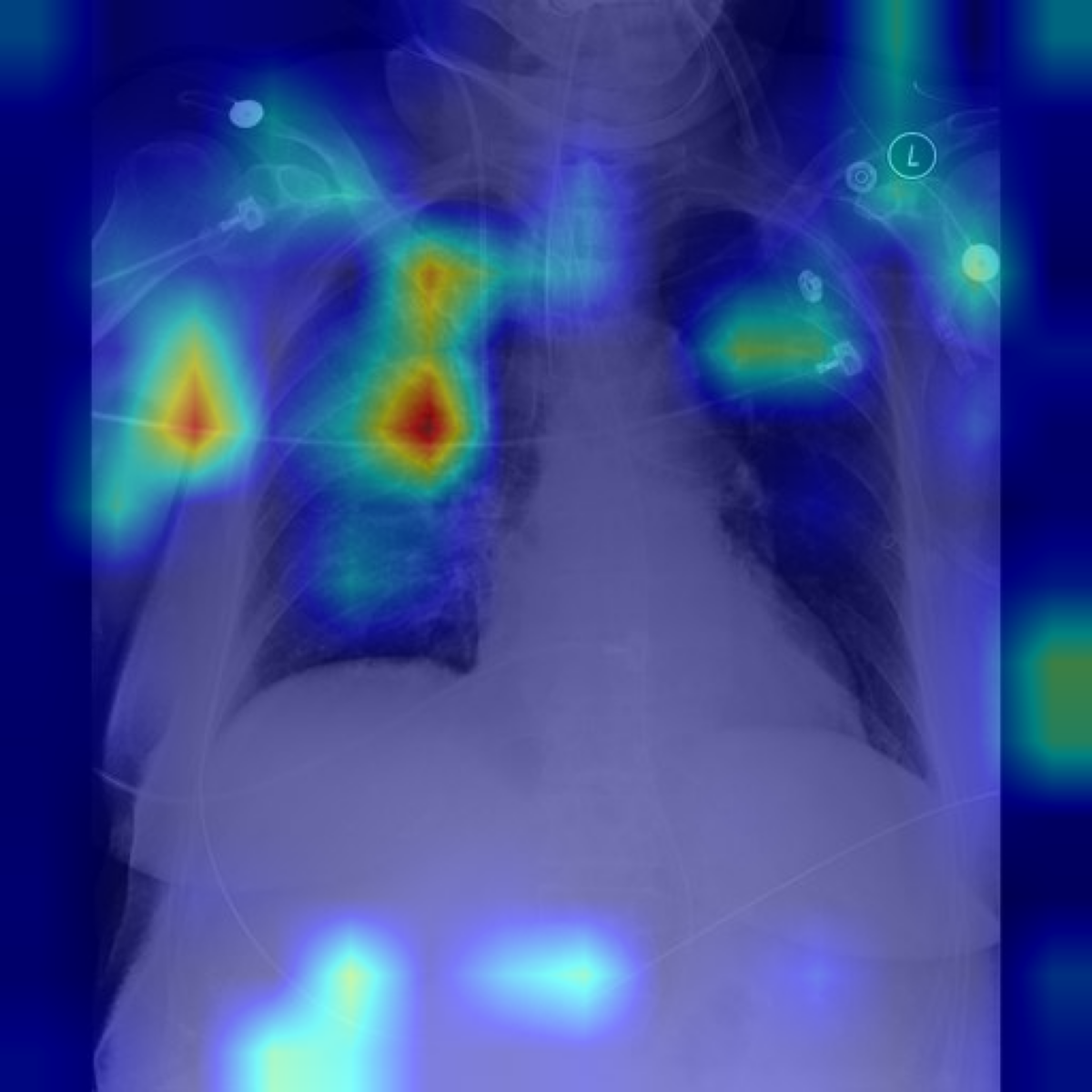}} \\[0.2cm]
& \textit{Endotracheal Tube} & \textit{Right Upper} \\[0.2cm]
\end{tabular}
} \\
\hline
\hline
\rowcolor{blue!10}
\textbf{Ground Truth} & \textbf{Generated Report} \\
\hline
The patient is status post \underline{median sternotomy} and aortic valve replacement. Heart size is moderately enlarged but unchanged. The aorta is tortuous and calcified. There is mild interstitial pulmonary edema, relatively unchanged. At least small \underline{bilateral pleural} effusions are present. Bibasilar airspace opacities may reflect compressive atelectasis. There is no pneumothorax. There is evidence of prior vertebroplasty at the thoracolumbar junction.
&
The patient is status post \colorbox{blue!20}{\textcolor{blue!90}{\textbf{median sternotomy}}} and CABG. The heart size remains mildly enlarged. The aorta is tortuous and diffusely calcified. There is mild pulmonary edema, worse in the interval, with small \colorbox{blue!20}{\textcolor{blue!90}{\textbf{bilateral pleural}}} effusions. Bibasilar airspace opacities likely reflect atelectasis. No pneumothorax is identified. There are no acute osseous abnormalities. \\
\hline
\multicolumn{2}{|c|}{
\cellcolor{blue!5}
\begin{tabular}{c c@{\hspace{0.6cm}}c}
\textbf{Original Image} & \multicolumn{2}{c}{\textbf{Grad-ECLIP Activation Maps}} \\
\rule{0pt}{0.5cm}
\fbox{\includegraphics[width=3.5cm]{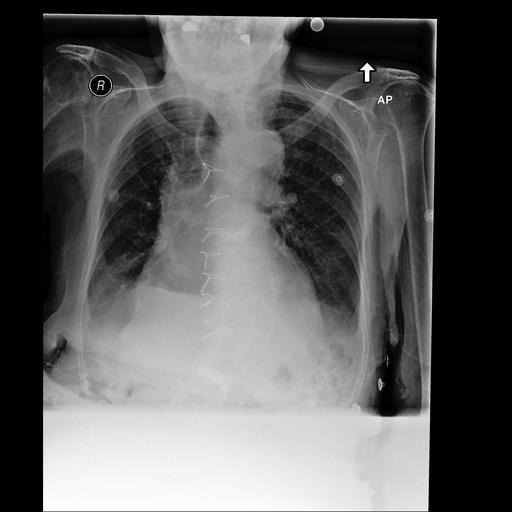}} &
\fbox{\includegraphics[width=3.5cm]{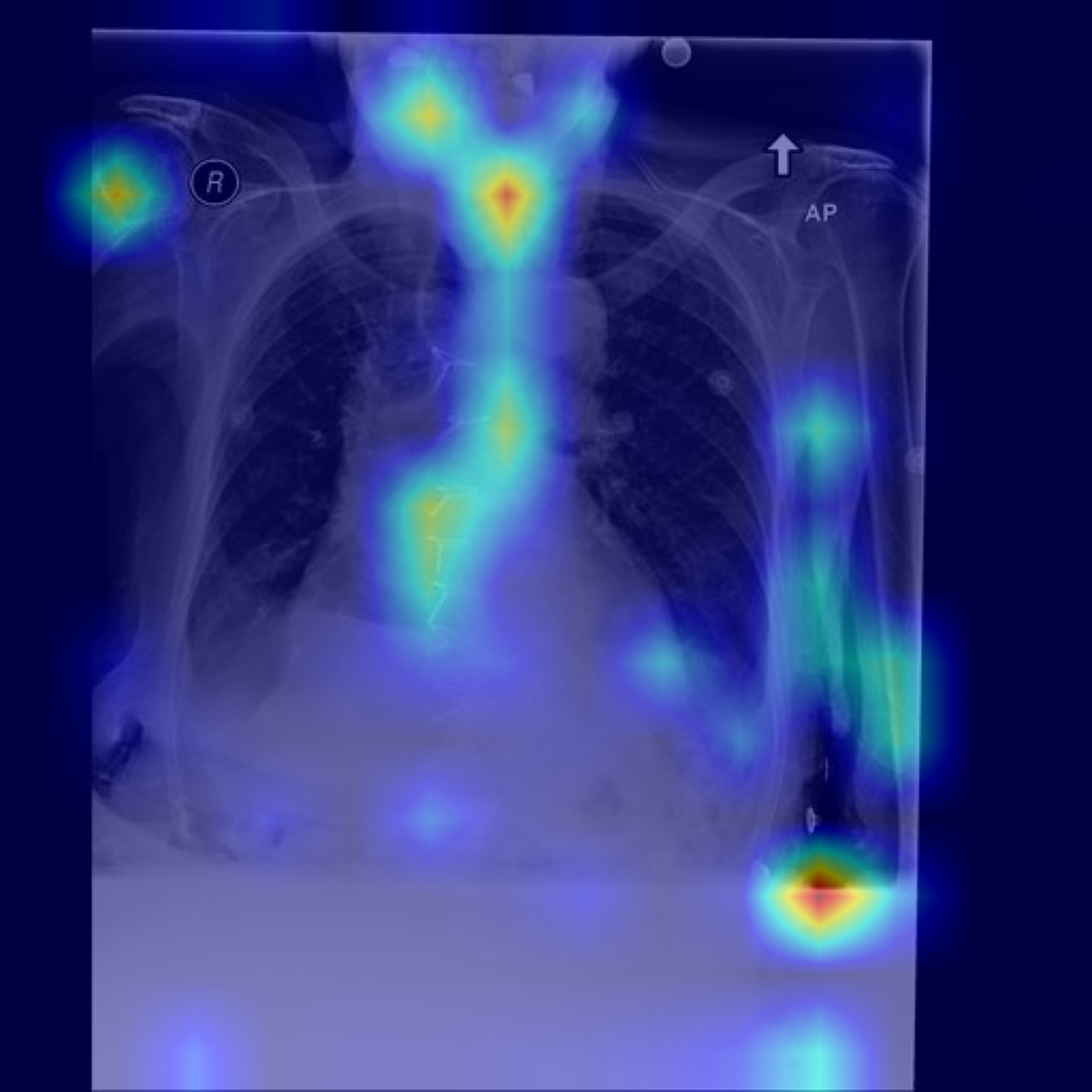}} &
\fbox{\includegraphics[width=3.5cm]{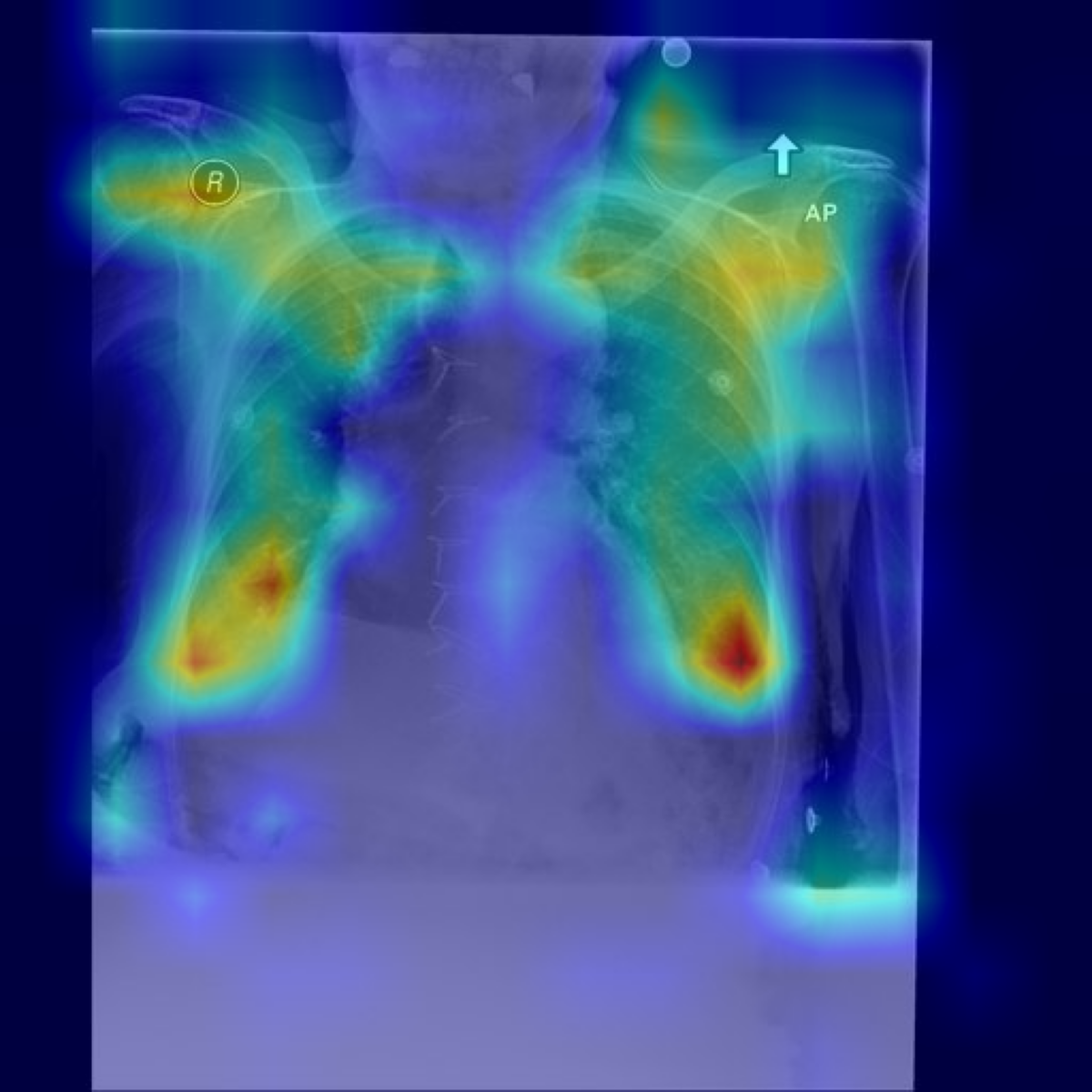}} \\[0.2cm]
& \textit{Median Sternotomy} & \textit{Bilateral Pleural} \\[0.2cm]
\end{tabular}
} \\
\hline
\end{tabular}
}
\end{table*}

For each case, we select SpLiCE concepts that appear in the generated report (e.g., \textit{endotracheal tube, right upper, median sternotomy, bilateral pleural}). We apply Grad-ECLIP~\cite{pmlr-v235-zhao24p} to compute the gradient of the image–text similarity score with respect to the visual features used by the LLaVA-Med vision encoder. The resulting relevance map is then upsampled and overlaid on the chest X-ray as a heat map, yielding a concept-specific visualization of which regions support that term.
\ModelName-generated reports and their corresponding concept heat maps are illustrated in \cref{tab:gradeclip_examples} for two representative MIMIC-CXR cases, one in the \emph{Zero-shot} regime and one after \emph{SFT}. In each example, the left column shows the reference report, while the right column shows the \ModelName output with the relevant concept tokens highlighted. Below these reports, we display the original image together with Grad-ECLIP maps for two selected concepts.
In the \emph{Zero-shot} example, concepts such as endotracheal tube and right upper are both present in the generated report and exhibit Grad-ECLIP activations that concentrate along the tracheal tube and in the right upper lung zone, respectively. 
In the \emph{SFT} example, the concept median sternotomy yields strong activation along the midline sternal wires and retrosternal region, while bilateral pleural produces strongest activation in the lower lung zones near the right and left costophrenic angles, where pleural effusions typically accumulate. 
Taken together, these patterns are consistent with standard radiological practice and suggest that, when a concept appears in the report, it is usually backed by visually plausible evidence in the corresponding image regions.

These concept-specific maps serve two complementary purposes. 
First, they allow clinicians to verify that SpLiCE-derived concepts, when injected into the prompt, are grounded in image regions that are consistent with their radiological meaning, rather than reflecting arbitrary directions in embedding space. Second, they expose failure modes in a transparent way: if a concept is mentioned in the report but its Grad-ECLIP map does not align with plausible anatomy, this discrepancy can be inspected, questioned, and potentially used to flag low-trust outputs.
At the same time, this analysis remains focused on the vision component: Grad-ECLIP explains how visual encoders support individual concepts, but does not by itself reveal how the LLM composes multiple concepts into full sentences. Extending concept-level tracing to the language component remains an important direction for future work toward fully interpretable medical VLMs.


\section{Conclusion}\label{sec6}

This work addressed two major barriers to the clinical deployment of vision–language models in radiology: limited interpretability and susceptibility to hallucinations. We introduced \ModelName, a unified framework that integrates concept decomposition with multimodal RAG to jointly enhance transparency and factual accuracy in radiology report generation. We further established a comprehensive evaluation protocol that compares multiple prompting strategies across different VLM architectures, retrieval configurations, and datasets, using both lexical similarity and clinically oriented correctness metrics. 
Across experiments on MIMIC-CXR and IU X-ray, our results indicate that interpretable visual concepts can improve factual grounding and concept-level transparency simultaneously, challenging the commonly assumed trade-off between interpretability and performance in medical AI.
From a clinical perspective, the proposed framework offers a practical way to present AI-generated draft reports together with explicit visual concepts and retrieved reference cases, potentially facilitating more efficient review while preserving radiologists’ ability to verify how findings in the image relate to the generated text.

Despite these advances, several limitations warrant further investigation. The overall effectiveness of the framework depends critically on the quality of the SpLiCE decomposition: if the underlying CLIP encoders are not sufficiently aligned with domain-specific semantics, the extracted concepts may be noisy or incomplete. Improving this alignment, for example through domain-adaptive pretraining or architectural refinements, is an important direction for future work. Moreover, interpretability in the current pipeline is primarily concentrated in the vision encoder and concept layer, while the language model is influenced only indirectly via prompt conditioning. Future research should explore mechanisms that more directly constrain or regularize token probabilities during generation, extending interpretability to the full model.
Finally, our experiments focused on an LLM backbone such as Mistral-7B; however, recent progress on smaller language models with competitive performance and lower computational cost suggests a promising avenue for achieving finer-grained control over interpretability and deployment in resource-constrained clinical settings.

By showing that concept-level interpretability can enhance rather than undermine factual accuracy, this work provides empirical support for the development of transparent and accurate VLMs for radiology. The modular design of \ModelName enables targeted optimization of individual components and offers a general methodological template that can be extended beyond chest X-ray analysis to other medical imaging domains where visual interpretation and textual reporting are required, provided that suitable domain-specific concept vocabularies and retrieval corpora are available.

\section{Acknoweldgments}
Marco Salmè is a Ph.D. student enrolled in the National Ph.D. in Artificial Intelligence, XXXIX cycle, course on Health and Life Sciences, organized by Università Campus Bio-Medico di Roma. This work was partially funded by: i) Università Campus Bio-Medico di Roma under the program “University Strategic Projects” within the project “AI-powered Digital Twin for next-generation lung cancEr cAre (IDEA)”; ii) PNRR MUR project PE0000013-FAIR. iii) Cancerforskningsfonden Norrland project MP23-1122; iv) Kempe Foundation project JCSMK24-0094. Resources are provided by the National Academic Infrastructure for Supercomputing
in Sweden (NAISS) and the Swedish National Infrastructure for Computing (SNIC) at Alvis @ C3SE, partially funded by the Swedish Research Council through grant agreements no. 2022-06725 and no. 2018-05973.

\bibliography{biblio}

\begin{appendices}

\section{Visual Concept Extraction}\label{secA}

For extracting interpretable visual concepts from CLIP embeddings, we employ Sparse Linear Concept Embeddings (SpLiCE)~\cite{bhalla2024interpreting}.
Given the CLIP visual embedding $\mathbf{v} \in \mathbb{R}^d$ extracted from the input image $\mathbf{I}$, SpLiCE approximates this embedding as a sparse linear combination of concept embeddings drawn from a learned vocabulary.

The concept vocabulary construction begins with a set of $m$ medical concepts $\mathbf{Q} = {[q_1, q_2, \dots, q_m]}$ encompassing radiological findings, anatomical structures, and pathological patterns relevant to chest radiograph interpretation. 
Each concept $q_j \in \mathbf{Q}$ is encoded through the CLIP text encoder $\mathrm{E}_{\text{txt}}$ to obtain its embedding
$\mathbf{c}_j = \mathrm{E}_{\text{txt}}(q_j) \in \mathbb{R}^d$,
and we collect these embeddings into the matrix $\mathbf{C} = [\mathbf{c}_1, \mathbf{c}_2, \dots, \mathbf{c}_m] \in \mathbb{R}^{d \times m}$.
We denote by $\sigma(\mathbf{x}) = \mathbf{x} / \|\mathbf{x}\|_2$ the normalization operator.
Let $\boldsymbol{\mu}_{\text{c}} \in \mathbb{R}^d$ be the mean concept embedding computed over $\mathbf{C}$.
The centered and normalized vocabulary is then given by $\tilde{\mathbf{c}}_j = \sigma(\mathbf{c}_j - \boldsymbol{\mu}_{\text{c}})$,
and $\tilde{\mathbf{C}} = [\tilde{\mathbf{c}}_1, \tilde{\mathbf{c}}_2, \dots, \tilde{\mathbf{c}}_m] \in \mathbb{R}^{d \times m}$,
where each element represents a centered and normalized concept embedding.
Analogously, we center and normalize the image embedding.
Let $\boldsymbol{\mu}_{\text{img}} \in \mathbb{R}^d$ denote the mean CLIP image embedding computed over the training corpus. 
The centered and normalized image embedding is then given by
$\tilde{\mathbf{v}} = \sigma(\mathbf{v} - \boldsymbol{\mu}_{\text{img}})$.
The sparse decomposition is formulated as an optimization problem that
balances reconstruction accuracy against sparsity:
\begin{equation}
\boldsymbol{\alpha}^* = \arg\min_{\boldsymbol{\alpha} \geq 0} \|\tilde{\mathbf{C}}\boldsymbol{\alpha} - \tilde{\mathbf{v}}\|_2^2 + 2\lambda\|\boldsymbol{\alpha}\|_1
\label{eq:lasso_opt}
\end{equation}
where $\boldsymbol{\alpha} = [\alpha_1, \alpha_2, \dots, \alpha_m]\in \mathbb{R}_{\ge 0}^m$ represents the coefficient vector encoding the contribution of each concept. The first term enforces reconstruction fidelity in the centered embedding space, while the second term promotes sparsity with the regularization parameter $\lambda > 0$ controlling the trade-off.
The solution to this optimization yields a sparse coefficient vector $\boldsymbol{\alpha}^* = [\alpha_1^*, \dots, \alpha_m^*] \in \mathbb{R}_{\ge 0}^m$, where only a small subset of entries are non-zero.
To obtain the final set of concept keywords, we rank concepts by their corresponding coefficient magnitudes and select the top-$\tau$ concepts. 
These selected concepts correspond to interpretable keywords $\boldsymbol{\Omega}$ that represent clinically relevant visual features present in the input image. The resulting keyword set $\boldsymbol{\Omega}$ provides transparency into which aspects of the image inform the subsequent RRG process and serves as the concept component for the prompt augmentation.

\subsection{Vocabulary Construction and Hyperparameter Choice}\label{secA1}
The vocabulary was constructed by extracting the most frequent bigrams from the training corpus, with systematic exclusion of English stopwords and common medical acronyms. 
The decision to focus on bigrams rather than unigrams is motivated by the inherent compositional nature of radiological terminology. 
Many clinically meaningful concepts emerge only through the combination of 
terms. For instance, ``pleural effusion'' and ``cardiomediastinal silhouette'' convey precise diagnostic information that cannot be adequately represented by their constituent words in isolation. 
All extracted terms underwent lemmatization to ensure morphological normalization and direct compatibility with downstream textual prompts fed to the LLM.

Having established the vocabulary construction methodology, we conducted systematic ablation studies to determine optimal parameters for the SpLiCE decomposition. The optimization process involved joint exploration of three critical hyperparameters: vocabulary size, L1 regularization strength ($\lambda$), and the number of top-ranked concepts ($\tau$) selected for each image. 
Our objective was to identify a configuration that simultaneously satisfies three competing criteria: (i) precision, (ii) cosine similarity and (iii) sparsity.
Precision, defined as the fraction of extracted SpLiCE concepts that appear in the corresponding ground truth radiology report, serves as our primary quality metric. 
This measure directly quantifies the clinical relevance and factual accuracy of the extracted concepts. 
High precision is essential in our framework, as these concepts are incorporated into retrieval-augmented prompts that guide report generation. 
Cosine similarity between the original CLIP image embedding and its sparse reconstruction quantifies the fidelity of the decomposition. 
This metric ensures that the dimensionality reduction and sparsification process preserves the essential visual information encoded in the original representation.
Sparsity, measured as the average number of non-zero coefficients per image, reflects the interpretability-informativeness trade-off. 
Excessive sparsity may omit clinically relevant concepts, while insufficient sparsity produces verbose, representations that overwhelm the downstream LLM with redundant information.

\begin{figure}[h]
    \centering
    \includegraphics[width=\textwidth]{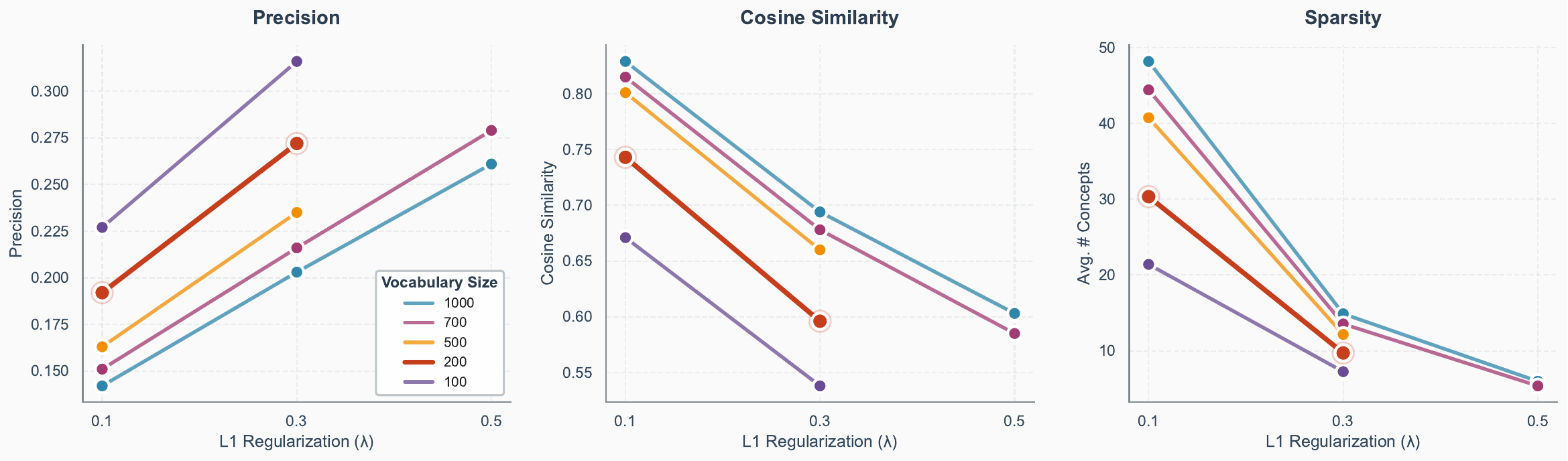}
    \caption{
    SpLiCE performance across vocabulary sizes and $\lambda$ values. 
    The figure reports three complementary metrics: 
    \textbf{Left}: precision of extracted concepts; 
    \textbf{Center}: cosine similarity between the original CLIP embedding and its sparse reconstruction; 
    \textbf{Right}: average number of active concepts (sparsity).
    The results highlight the trade-off between fidelity, interpretability, and terminological precision.
    }
    \label{fig:splice_optimization}
\end{figure}

~\cref{fig:splice_optimization} presents a comprehensive analysis across five vocabulary sizes, corresponding to the top 100, 200, 500, 700, and 1000 most frequent bigrams from the MIMIC-CXR training corpus, and three L1 penalty values ($\lambda \in \{0.1, 0.3, 0.5\}$). 
It should be noted that for vocabulary sizes below 500, results at $\lambda=0.5$ are unavailable due to over-regularization: the strong sparsity penalty caused certain images to yield entirely zero-valued weight vectors, indicating that no concept exceeded the activation threshold. 
This phenomenon confirms the theoretical prediction that excessively high $\lambda$ values can suppress all activations when the vocabulary-embedding alignment is insufficient.
Precision trends (left panel of ~\cref{fig:splice_optimization}) reveal a pronounced inverse relationship with vocabulary size. 
This systematic pattern reflects a fundamental trade-off between lexical coverage and terminological precision: smaller vocabularies, comprising only the most frequently occurring clinical terms, naturally align with the standardized terminology that radiologists consistently employ in their reports, whereas larger vocabularies introduce lower-frequency terms that, despite exhibiting visual correlation with CLIP embeddings, often represent rare synonyms, anatomical descriptors, or overly specific variants absent from actual ground truth reports. Within each vocabulary configuration, increasing $\lambda$ consistently enhances precision by enforcing greater sparsity, thereby selecting only the most strongly activated concepts while reducing false positive extractions.
Reconstruction fidelity (central panel of ~\cref{fig:splice_optimization}) exhibits the opposite trend: cosine similarity increases monotonically with vocabulary size, ranging from 0.671 for the 100-bigram vocabulary at $\lambda=0.1$ to 0.829 for the 1000-bigram vocabulary at the same regularization strength. This observation is expected, as larger dictionaries provide greater representational capacity to approximate the original high-dimensional CLIP embedding through linear combinations. 
However, this improved reconstruction comes at the documented cost of reduced precision.
Sparsity characteristics (right panel of ~\cref{fig:splice_optimization}) demonstrate that the average number of active concepts decreases both with vocabulary size reduction and $\lambda$ increment. 
Large vocabularies with minimal regularization, such as the 1000-bigram vocabulary at $\lambda=0.1$, produce approximately 48 non-zero coefficients per image, a density incompatible with interpretable prompting and efficient retrieval. Conversely, aggressive sparsification achieved through the 100-bigram vocabulary at $\lambda=0.3$ yields only 6-7 active concepts, representing a focused subset that provides sufficient contextual information to guide generation while avoiding overwhelming the language model with excessive detail.
In selecting the final configuration, we assigned priority to precision maximization, recognizing that extracted concepts directly influence the factual accuracy of generated reports, thereby identifying a vocabulary size of 100 or 200 as the most promising candidates, both achieving precision exceeding 0.27.

\begin{figure}[h]
    \centering
    \includegraphics[width=0.75\textwidth]{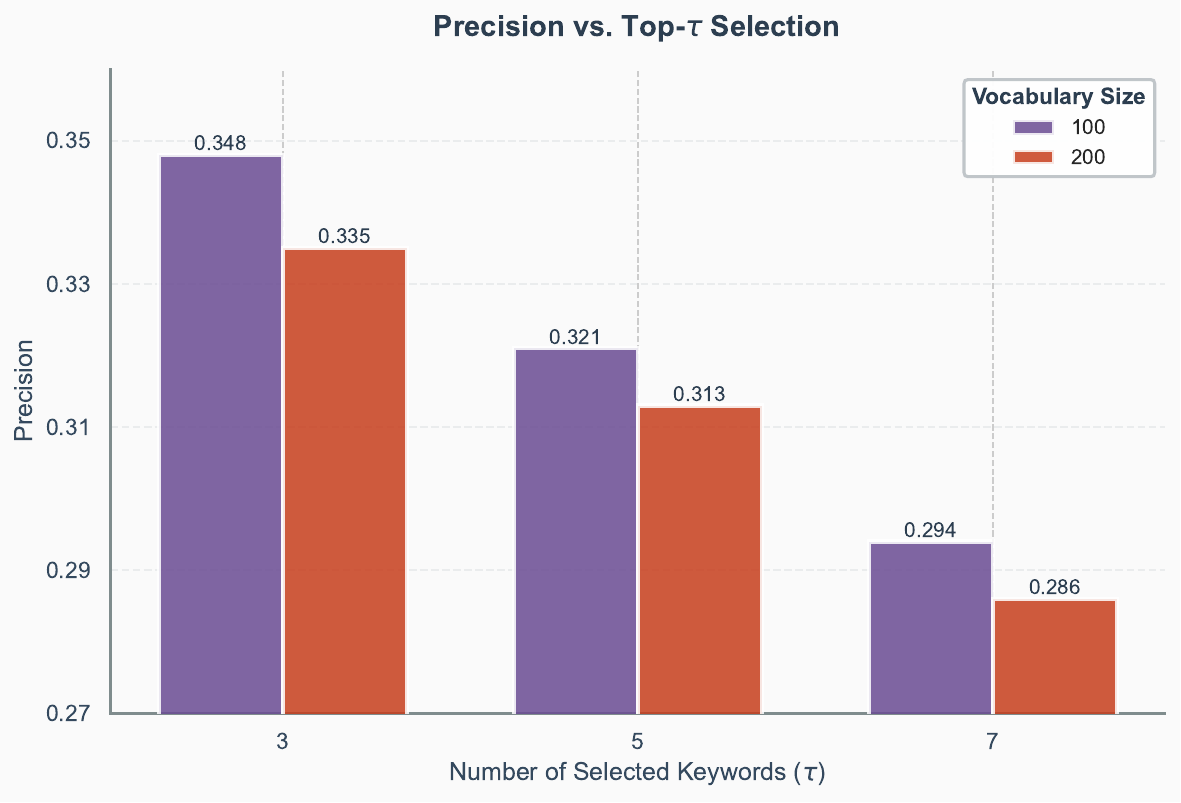}
    \caption{
    Precision as a function of the number of selected concepts $\tau \in \{3, 5, 7\}$ 
    for both the 100- and 200-bigram vocabularies. 
    Precision decreases monotonically with increasing $\tau$, reflecting the 
    progressive inclusion of lower-ranked concepts with weaker activation strengths. 
    The 100-bigram vocabulary maintains a consistent advantage across all $\tau$ values, 
    indicating that the highest-confidence concepts are highly informative for guiding 
    retrieval-augmented report generation.
    }
    \label{fig:topk_precision}
\end{figure}

To further refine this selection and ensure consistency across the dataset, we conducted an additional experiment evaluating the impact of selecting a fixed number $\tau$ of top-ranked concepts per image, with $\tau \in [3, 5, 7]$. 
This analysis served two purposes: first, to assess whether further reduction in concept count would significantly degrade precision; and second, to standardize the input representation such that all images contribute exactly the same number of keywords to the retrieval-augmented prompts, facilitating uniform processing by the downstream LLM. 
Results are presented in ~\cref{fig:topk_precision}, where precision decreases monotonically with increasing $\tau$ for both vocabularies, consistent with the progressive inclusion of concepts exhibiting lower activation coefficients.
Notably, the precision advantage of 100-bigram vocabulary diminishes as $\tau$ increases, from a 3.9\% margin at $\tau=3$ to 2.8\% at $\tau=7$, suggesting that the highest-ranked concepts exhibit comparable accuracy across both vocabularies, with divergence manifesting primarily in lower-ranked selections.
We adopt $\tau=5$ with 200-bigram vocabulary as our definitive configuration (precision=0.313). 
While the vocabulary size of 100 offers marginally superior precision, the 200-bigram vocabulary provides substantially broader lexical coverage, encompassing twice the vocabulary size, with only modest precision degradation.
This configuration embodies a principled balance between factual grounding through high precision and expressive capacity through adequate lexical diversity, optimizing the extracted concepts for their ultimate role in guiding accurate and comprehensive RRG via retrieval-augmented prompting.

\section{Training Details}
\label{app:training}

\paragraph*{CXR-CLIP adaptation.} For the IU X-ray experiments, we adapt CXR-CLIP using LoRA~\cite{hu2022lora}. 
The LoRA modules are applied to the last stage of the Swin vision encoder and to the final BERT text encoder layer, with rank $r=8$, scaling parameter $\alpha=16$, and dropout rate $0.1$.
Training is performed for ten epochs using the AdamW optimizer, with a learning rate of $5 \times 10^{-5}$ and a weight decay of $0.01$.

\paragraph*{Projection module $\Phi_{\text{CLIP}}$.} The projection module $\Phi_{\text{CLIP}}$ is implemented as a single linear layer that maps the 768-dimensional CLIP visual representation from the last vision transformer block (before the standard CLIP projection head) to the LLM embedding dimension.
As an alignment step, $\Phi_{\text{CLIP}}$ is initially trained for one epoch with both the CLIP encoder and the LLM frozen, using a learning rate of $1 \times 10^{-3}$, a warmup ratio of $0.03$, cosine learning-rate scheduling, and a batch size of $16$.

\paragraph*{Supervised Fine-Tuning (SFT).} For SFT of the LLM, we again employ LoRA with rank $r=64$, scaling parameter $\alpha=16$, and dropout rate $0.05$.
Fine-tuning is carried out for three epochs with a batch size of $16$, a learning rate of $1 \times 10^{-4}$, a warmup ratio of $0.03$, and cosine learning-rate scheduling, while keeping all visual encoders frozen.

\end{appendices}

\end{document}